\documentclass{article}

\usepackage[preprint]{neurips_2026}

\usepackage[utf8]{inputenc}
\usepackage[T1]{fontenc}
\usepackage{hyperref}
\usepackage{url}
\usepackage{booktabs}
\usepackage{amsfonts}
\usepackage{amssymb}
\usepackage{amsmath}
\usepackage{nicefrac}
\usepackage{microtype}
\usepackage{xcolor}
\usepackage{graphicx}
\usepackage{multirow}
\usepackage{array}
\usepackage{enumitem}
\usepackage{subcaption}
\usepackage{algorithm}
\usepackage{algpseudocode}
\usepackage{tikz}
\usetikzlibrary{arrows.meta, positioning, fit, backgrounds, shapes.geometric, calc}
\usepackage{pifont}
\newcommand{\cmark}{\ding{51}}
\newcommand{\xmark}{\ding{55}}

\title{ReFlect: An Effective Harness System\\for Complex Long-Horizon LLM Reasoning}

\author{%
  Fan Huang \\
  Indiana University Bloomington \\
  \texttt{huangfan@acm.org}
}

\begin{document}

\maketitle

\begin{abstract}
Current reasoning paradigms for LLMs include chain-of-thought, ReAct, and post-hoc self-critique. These paradigms rely on two assumptions that fail on long-horizon, multi-stage tasks. As a result, errors accumulate silently across reasoning steps, leaving an open question: can a reasoning system effectively detect and recover from its own failures? We present ReFlect, a \emph{harness} system for LLM reasoning that creates standalone error detection and recovery logic as a deterministic wrapper around the model. Controlled experiments across 6 reasoning domains show that prompt-level self-critique produces formulaic templates that flag no issues in 90 of 100 audited reflection blocks, and the investigated LLMs wrongly accept a wrong answer in at least 76\% of cases. Our ReFlect harness achieves task success rates ranging from 41\% on gpt-4o-mini to 56\% on Claude Sonnet 4.5 across six models spanning small and frontier scale, with per-model gains over Direct CoT ranging from $+7$~pp on Qwen2.5-72B to $+29$~pp on Claude Sonnet 4.5, and additionally raises SWE-bench patch-structural quality from 0\% (Direct CoT) to between 82\% (Qwen2.5-72B) and 87\% (GPT-4o). Notably, the harness gain is inversely proportional to the model's Direct CoT task success rate (the fitted slope is $-1.69$ with $r=-0.76$): each pp lost in baseline success rate is mechanically recovered by 1.69~pp of harness gain. We spot that adding structured reasoning state and operators yields only 15.0--18.7\% pair-mean on Llama-3.3-70B and Qwen2.5-72B because models at this scale cannot reliably populate the state its operators require. ReFlect is model-agnostic, training-free, and operates entirely at inference time.
\end{abstract}

\section{Introduction}
\label{sec:intro}

Large language models (LLMs) are increasingly deployed on complex, long-horizon, multi-stage reasoning tasks (multi-file code engineering~\citep{jimenez2024swebench}, multi-document scientific synthesis~\citep{dasigi2021qasper}, olympiad-level mathematics~\citep{aime2024}, and action-grounded household planning~\citep{shridhar2020alfred}), building on advances in step-by-step reasoning~\citep{wei2022chain}, interleaved reasoning and acting~\citep{yao2022react}, and deliberative search~\citep{yao2024tree}. Unlike single-pass question answering, these tasks demand not just competent local reasoning but the ability to detect when reasoning has gone wrong, structurally validate intermediate outputs, and recover deterministically (each failure maps to a predetermined recovery action, e.g., retry with a stricter format or fall back to a different tool).

For these complex reasoning tasks, most existing agentic LLM reasoning systems exhibit recurring failures: errors accumulate silently across the trajectory~\citep{arbuzov2025errors,sinha2026longhorizon}, models exhibit a self-correction blind spot that prevents them from detecting their own errors~\citep{tsui2025blindspot}, and no deterministic recovery procedure exists once a problem arises. Recent surveys of LLM-agent evaluation and benchmarking~\citep{mohammadi2025agentsurvey} identify reliability and recovery as central deployment obstacles, producing unpredictable end-to-end failures whose root cause is hard to attribute or fix.

Existing paradigms address this at the prompt level. Chain-of-thought (CoT)~\citep{wei2022chain}, ReAct~\citep{yao2022react}, Self-Refine~\citep{madaan2023selfrefine}, Reflexion~\citep{shinn2023reflexion}, CRITIC~\citep{gou2024critic}, IterResearch~\citep{chen2026iterresearch}, and tool-use frameworks (Toolformer~\citep{schick2023toolformer}, ART~\citep{paranjape2023art}, Gorilla~\citep{patil2024gorilla}) all locate detection-and-recovery logic inside the prompt or the model's free-text trajectory. They share two implicit assumptions that become untenable as task complexity grows: that local progress implies global progress, and that models can meaningfully self-correct by re-reading their own outputs. Yet recent work shows that current LLMs cannot reliably self-correct via LLM-judged critique~\citep{huang2024selfcorrection,pan2024selfcorrection}, and that interactive observation~\citep{yao2022react} or post-hoc reflection~\citep{madaan2023selfrefine,shinn2023reflexion} does not address the underlying detection problem.

We propose ReFlect, a harness that wraps the model with deterministic error-detection and recovery logic. A motivating pilot (\S\ref{sec:motivation}) first establishes that prompt-level self-critique does not suffice on long-horizon multi-stage reasoning; we then evaluate two ReFlect instantiations against three questions. RQ1 (heavyweight harness): does a design with explicit structured state, four reflective operators, and a regime controller compensate for missing model capability at 70B? RQ2 (lightweight harness): does a deterministic shape-routed design compensate for missing capability, with gain scaling inversely with base capability? RQ3 (what carries the gain): which of five standard primitives — structured-state operators, inspection calls, tool dispatch, structural validators, or computation routing — actually deliver gain on Llama-3.3-70B and Qwen2.5-72B?

We contribute from five aspects. (i) A harness framework with heavyweight (structured state, operators) and lightweight (shape routing, tool dispatch) Level-3 instantiations. (ii) RQ1 maps a base-capability prerequisite: structured-state heavyweight yields 15.0--18.7\% pair-mean on the 70B pair because state cannot be reliably populated; deterministic Python routing in the same family lifts pair-mean to 28.0\%. (iii) RQ2 establishes a \emph{capability-compensation} effect: harness gain is inversely proportional to Direct CoT accuracy (slope $-1.69$, $r=-0.76$), implying outsized benefits for cheap-model deployment. (iv) Beyond-accuracy evidence on convergence, stability, and token efficiency, with per-tool decomposition isolating what carries the gain. (v) Pilot $+$ 28-variant ablation: 90/100 reflections are formulaic with $\leq$1.7\% course correction, and LLMs wrongly accept incorrect answers in $\geq$76\% of cases across all Level-2 verifiers.

\section{Related work}
\label{sec:related}

\paragraph{Single-pass and interactive reasoning.}
Chain-of-thought prompting~\citep{wei2022chain}, self-consistency~\citep{wang2023selfconsistency}, least-to-most prompting~\citep{zhou2023leasttomost}, and the training-time STaR variant~\citep{zelikman2022star} produce reasoning traces in a single forward pass. ReAct~\citep{yao2022react} grounds intermediate steps in environment observations; Tree of Thoughts~\citep{yao2024tree} adds branching search over reasoning paths. These methods improve local step quality but treat the token trajectory as the only working artifact: no state representation lives outside the model's natural-language trajectory, and the only available error-detection signal is what the model itself verbalizes.

\paragraph{LLM-judged self-correction.}
Self-Refine~\citep{madaan2023selfrefine} decouples generation from critique via separate LLM calls iterating to convergence. Reflexion~\citep{shinn2023reflexion} adds cross-episode memory using environment feedback (code tests, game scores) and retries the full episode with that context. CRITIC~\citep{gou2024critic} verifies via external tools but issues critique as natural-language text. In all three, the critique step is itself an LLM call reading and writing free-text outputs, with no deterministic check between generation and revision; recent evaluations~\citep{huang2024selfcorrection,pan2024selfcorrection} report that current models cannot reliably self-correct in this regime. Our pilot (\S\ref{sec:motivation}) reproduces this finding for inline mid-task critique on Llama-3.3-70B and Qwen2.5-72B.

\paragraph{Long-horizon state and tool dispatch.}
Both threads in this category leave a piece of agentic infrastructure inside the model. IterResearch~\citep{chen2026iterresearch} maintains a bounded workspace $(q, M_t, \{a_{t-1}, \mathit{TR}_{t-1}\})$ that the model rewrites each round, scaling interaction to 2{,}048 turns at constant context size; $M_t$ is unstructured natural-language text, a memory rather than a verifier, with no programmatic contradiction detection. Tool-use frameworks Toolformer~\citep{schick2023toolformer}, ART~\citep{paranjape2023art}, and Gorilla~\citep{patil2024gorilla} teach or prompt the model to invoke external tools, leaving the routing decision (which tool, when) inside the model's generation stream. We instead externalize both: a feature-based shape classifier dispatches each problem to a tool registry deterministically, per-tool format validators with retry handle malformed outputs, and the layer composes naturally with workspace-reconstruction substrates of the IterResearch~\citep{chen2026iterresearch} or CoALA~\citep{sumers2024cognitive} kind.

\section{A taxonomy of reasoning paradigms}
\label{sec:taxonomy}

The methods surveyed in \S\ref{sec:related} differ along three structural axes: where state lives, where error detection occurs, and what recovery action a failure triggers. Mapping methods onto these axes (Table~\ref{tab:paradigm-comparison}) shows that the existing literature clusters in a region we call \emph{LLM-judged self-correction}: the critique step is itself an LLM call reading and writing free-text outputs, with no deterministic check between generation and revision. We adopt the following four-tier shorthand throughout the paper.

\paragraph{Levels 0--2: prior approaches.}
\textbf{Level 0 (single-pass generation):} CoT and its inference-time variants. State is the token trajectory; there is no error detection or recovery. \textbf{Level 1 (interactive observation):} ReAct and Tree of Thoughts. Adds environment observation or search-tree branching, but state remains text-level and error detection is heuristic at best. \textbf{Level 2 (LLM-judged self-correction):} Self-Refine, Reflexion, CRITIC, IterResearch, and the inline mid-task self-critique we evaluate as \emph{Minimal ReFlect} (\S\ref{sec:motivation}). The shared structural property at Level 2 is that the critique step is itself an LLM call reading and writing free-text; recovery is bounded by what the model can verbalize and re-generate, with no deterministic verifier in the loop.

\paragraph{Level 3: structural harnessing (this work).}
Detection and recovery live outside the model. A deterministic shape classifier dispatches each problem to a specialized tool, format validators mechanically reject malformed outputs, and a retry-as-code policy triggers either a stricter regeneration or a fall-back tool. The model is invoked only inside structurally-bounded slots whose validity the harness can mechanically check. ReFlect admits two instantiations: a \emph{lightweight} design with a shape classifier and tool registry (Figure~\ref{fig:architecture}; \S\ref{sec:lightweight}, our headline result) and a \emph{heavyweight} design with structured state, four operators, and a regime-aware controller (Algorithm~\ref{alg:reflect}; \S\ref{sec:heavyweight}, evaluated across the full pilot-study progression at 70B and shown to plateau without strict regression).

\begin{table}[t]
\caption{Inference-time reasoning paradigms compared along three structural axes: where state lives, where error detection occurs, and what recovery action is taken on failure. Methods cluster at Level 2 (state and critique both inside the LLM); ReFlect occupies Level 3 by externalizing both.}
\label{tab:paradigm-comparison}
\centering
\small
\setlength{\tabcolsep}{4pt}
\begin{tabular}{@{}cl>{\raggedright\arraybackslash}p{2.4cm}>{\raggedright\arraybackslash}p{2.9cm}>{\raggedright\arraybackslash}p{3.2cm}@{}}
\toprule
\textbf{Lvl} & \textbf{Method} & \textbf{State} & \textbf{Error detection} & \textbf{Recovery on failure} \\
\midrule
0 & CoT~\citep{wei2022chain} & Token trajectory & None & None \\
0 & Self-cons.~\citep{wang2023selfconsistency} & $K$ trajectories & Vote disagreement & Majority vote \\
1 & ReAct~\citep{yao2022react} & Trajectory $+$ obs. & Environment grounding & None \\
1 & ToT~\citep{yao2024tree} & Search tree (text) & Heuristic value & Branch search \\
2 & Self-Refine~\citep{madaan2023selfrefine} & Output text & LLM critic post-output & Regenerate \\
2 & Reflexion~\citep{shinn2023reflexion} & Cross-episode mem. & Environment feedback & Retry from scratch \\
2 & CRITIC~\citep{gou2024critic} & Output text & External tool & Edit \\
2 & IterResearch~\citep{chen2026iterresearch} & Free-text workspace & LLM-judged self-revision & Next-round revision \\
2 & Min.\ ReFlect & Token trajectory & Inline LLM checklist & None (narrate only) \\
3 & \textbf{ReFlect} & Tool-local; opt.\ structured & Deterministic format validator & Retry-as-code; fall-back tool \\
\bottomrule
\end{tabular}
\end{table}

\section{When Prompt-Level Self-critique Fails?}
\label{sec:motivation}

This section addresses a motivating question that precedes the three RQs: \emph{does prompt-level self-critique suffice on long-horizon multi-stage reasoning?} We establish empirically that the simplest instantiation of inline self-critique (asking models to periodically pause and audit their own reasoning) fails systematically; this failure motivates the structured harness designs evaluated as RQ1 (heavyweight, \S\ref{sec:method}) and RQ2 (lightweight, \S\ref{sec:lightweight-section}).

\subsection{Pilot study: setup and results}
\label{sec:pilot-setup}

We run 360 controlled inferences crossing four factors: two \emph{models} (Qwen2.5-72B-Instruct~\citep{qwen2024qwen25}, Llama-3.3-70B-Instruct~\citep{llama2024llama33}; vLLM~\citep{kwon2023vllm}, $T{=}0.6$, top-$p{=}0.95$); three \emph{methods} (Direct LLM, ReAct, and Minimal ReFlect — ReAct plus a 5-point checklist over state, consistency, assumptions, direction, and decision, inserted every 3 steps in the same generation stream); six \emph{domains} (SWE-bench Lite~\citep{jimenez2024swebench}, QASPER~\citep{dasigi2021qasper}, ProofWriter depth-5~\citep{tafjord2021proofwriter}, AIME~\citep{aime2024}, ALFRED~\citep{shridhar2020alfred}, FinQA~\citep{chen2021finqa}); and 10 \emph{problems} per cell, drawn from the 50-per-domain benchmark used for the main experiments (\S\ref{sec:experiments}). The central finding: Minimal ReFlect never outperforms either baseline on any domain for either model (per-domain table in Appendix~\ref{app:pilot-results-table}). Despite more tokens and more structured trajectories, prompt-level self-critique does not translate into better answers.

\subsection{Root causes and the fundamental flaw}
\label{sec:root-causes}

Five root causes explain this failure (full detail in Appendix~\ref{app:pilot-rootcauses}; structural metrics in Appendix~\ref{app:pilot-structural}): repetition loops in 7\% (Qwen) and 23\% (Llama) of ReFlect runs; over 90\% of reflection blocks are formulaic ``no-issues, on-track'' templates with zero corrective signal; reflection changes an answer in 1/60 Qwen and 0/60 Llama runs (Wilson 95\% CI $\leq 8.9\%$); reflection overhead truncates 25\% of Llama runs; and Llama largely ignores the structured reflection format. All five trace to a single architectural mistake: the model is asked to be both thinker and auditor in the same generation stream — no explicit state to inspect, no separation between reasoning and meta-reasoning, no harness deciding when to intervene (it fires mechanically every $N$ steps), and no mechanism to modify state (the model only narrates). The conclusion is not ``self-critique doesn't help'' but \emph{prompting for self-critique $\neq$ structural harnessing}: without a wrapper outside the prompt (deterministic classification, specialized tools, format validators, retry-as-code), self-critique degenerates into narration.

\section{Experimental setup}
\label{sec:setup}
\label{sec:exp-setup}

We evaluate ReFlect across four experimental scopes — the motivating pilot plus one scope per RQ. This section consolidates domains, methods, backbones, serving infrastructure, and metrics so that the per-RQ sections (\S\ref{sec:method}, \S\ref{sec:lightweight-section}, \S\ref{sec:heavyweight-vs-lightweight}) can focus on framework design and per-RQ analysis.

\paragraph{Domains.}
Six reasoning domains span the structural shapes the harness routes between: SWE-bench Lite~\citep{jimenez2024swebench} (artifact generation, unified-diff format), QASPER~\citep{dasigi2021qasper} (evidence extraction from scientific papers), ProofWriter (depth-5)~\citep{tafjord2021proofwriter} (logical inference under closed-world rules), AIME 2022 to 2024~\citep{aime2024} (olympiad mathematics, symbolic), ALFRED~\citep{shridhar2020alfred} (procedural action-grounded planning), and FinQA~\citep{chen2021finqa} (tabular financial question answering). Each domain contributes 50 instances to the main grid (300 total); the motivating pilot draws 10 instances per (model, method, domain) cell from these same 300 (giving 360 controlled inferences across 2 models $\times$ 3 methods $\times$ 6 domains $\times$ 10 problems); the RQ1 heavyweight progression and the RQ3 ablation each use the full 300 per (model, variant) cell.

\paragraph{Methods compared.}
\textbf{Direct} (Level 0): single-pass step-by-step generation. \textbf{ReAct}~\citep{yao2022react} (Level 1): interleaved Thought$\to$Action$\to$Observation. \textbf{Self-Refine}~\citep{madaan2023selfrefine} (Level 2, 3 rounds): generate--critique--revise via separate LLM calls. \textbf{Reflexion}~\citep{shinn2023reflexion} (Level 2, 3 episodes): cross-episode memory plus environment feedback. \textbf{Full ReFlect} (Level 3 lightweight, \S\ref{sec:lightweight}): shape classifier $+$ tool registry. 

\paragraph{Backbones and serving infrastructure.}
The 70B pair (Llama-3.3-70B-Instruct~\citep{llama2024llama33}, Qwen2.5-72B-Instruct~\citep{qwen2024qwen25}) is shared across the motivating pilot, RQ1 heavyweight, and RQ3 ablation, served on vLLM~\citep{kwon2023vllm} (bf16, local GPUs, tensor-parallel across 4 GPUs, max-model-len 32{,}768). RQ2 lightweight serves the same 70B pair plus four additional models on remote APIs: Together.ai (Llama as the FP8 Turbo variant) and OpenRouter (Qwen) for the 70B pair, plus Claude Haiku 4.5 (\texttt{claude-haiku-4-5}) and Claude Sonnet 4.5 (\texttt{claude-sonnet-4-5}) via the Anthropic API and gpt-4o-mini and GPT-4o via the OpenAI API for the 6-model capability ladder. Serving details, FP8/bf16 reproducibility caveats, and per-tool sampling parameters are in Appendix~\ref{app:serving}.

\paragraph{Metrics.}
Domain-specific correctness uses test pass rate (SWE-bench), token-level F1 (QASPER), exact match (ProofWriter, AIME), action-sequence accuracy (ALFRED), and numerical accuracy (FinQA). SWE-bench is additionally reported under a tiered structural-quality scorer (0.0/0.3/0.6/1.0 for diff format $\to$ code-file targeting $\to$ Python AST validity; Appendix~\ref{app:swebench-scorer}). Beyond accuracy, lightweight-harness effectiveness (RQ2, \S\ref{sec:compute-matched}) is measured by convergence rate (Appendix~\ref{app:convergence}), token efficiency (Table~\ref{tab:compute-matched}), per-tool contribution (Appendix~\ref{app:per-tool}), verifier FP rate (Appendix~\ref{app:error-correction}), and stable-error recurrence (Appendix~\ref{app:repeated-error}).


\section{RQ1: Heavyweight Harness}
\label{sec:method}
\label{sec:heavyweight-section}


\subsection{Heavyweight harness instantiation}
\label{sec:heavyweight}

\begin{figure}[t]
\centering
\includegraphics[width=0.95\linewidth]{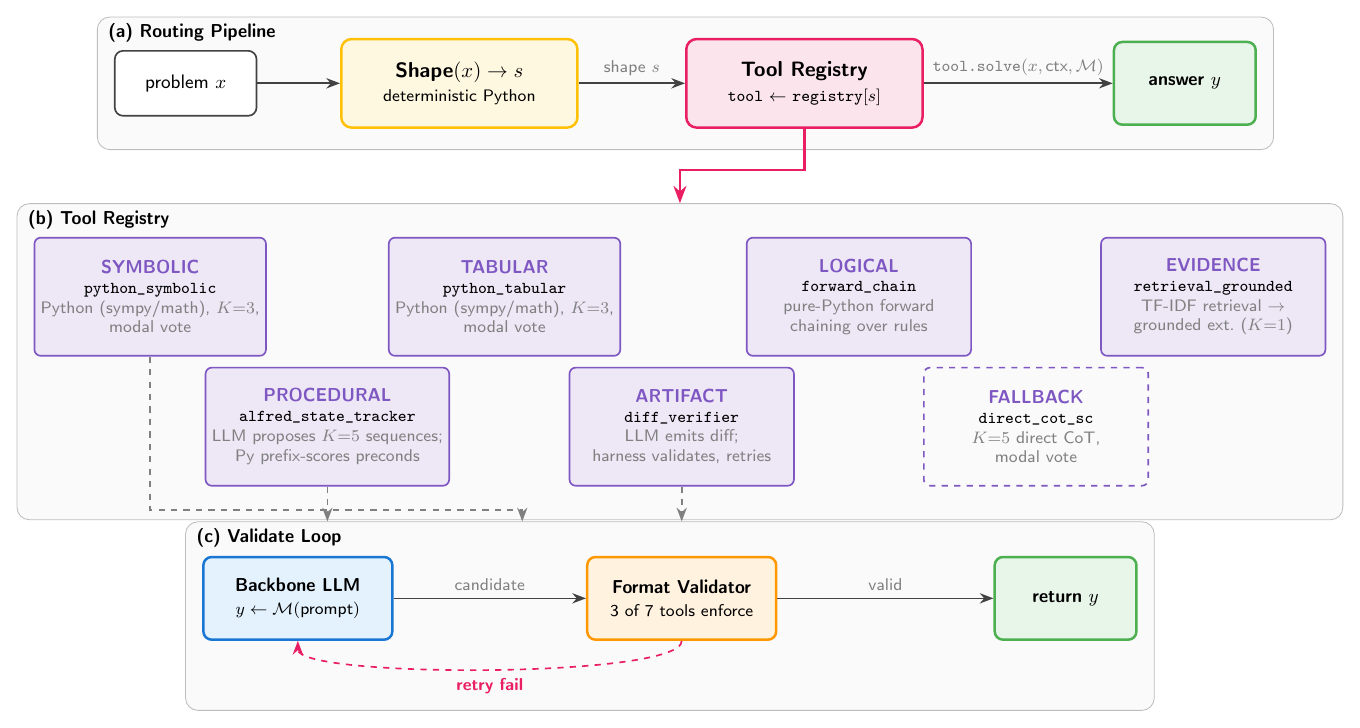}
\caption{Lightweight ReFlect architecture (the headline \emph{Full ReFlect} of \S\ref{sec:main-results}). \textbf{(a)} Routing pipeline: a deterministic-Python classifier $\textsc{Shape}: x \mapsto s$ dispatches each problem via the tool registry to a tool that handles it end-to-end through \texttt{tool.solve}$(x, \text{ctx}, \mathcal{M})$. \textbf{(b)} Tool registry contents: four generic shapes (\textsc{Symbolic}, \textsc{Tabular}, \textsc{Logical}, \textsc{Evidence}), two task-specific (\textsc{Procedural}, \textsc{Artifact}; added in Full ReFlect), plus \textsc{Fallback} (dashed). $K$ = independent LLM samples (modal-voted). \textbf{(c)} Validate Loop: the validator-with-retry pattern used by 3 of 7 tools (\textsc{Symbolic}, \textsc{Procedural}, \textsc{Artifact}); the other 4 execute via $K$-sample modal vote or deterministic computation.}
\label{fig:architecture}
\end{figure}

The heavyweight harness is the conceptually-natural Level-3 instantiation: a structured reasoning state $\mathcal{S} = (\mathcal{G}, \mathcal{A}, \mathcal{E}, \mathcal{D}, \mathcal{C}, \mathcal{T}, \mathcal{K}, r, u)$ (goal tree, assumptions with dependency cascade, sourced evidence, decisions, conflicts, compressed trajectory, checkpoints, regime, composite uncertainty); four reflective operators (\textsc{Inspect}, \textsc{Stabilize}, \textsc{Transform}, \textsc{Diversify}); and a controller that switches between five regimes (\textsc{Explore}, \textsc{Execute}, \textsc{Verify}, \textsc{Recover}, \textsc{Consolidate}). Full schema, operator definitions, controller policy, regime-transition rules, and detailed pseudocode (Algorithm~\ref{alg:reflect}) are in Appendix~\ref{app:heavyweight-detail}. We evaluated the design across five iterations on the 70B pair (Llama-3.3-70B and Qwen2.5-72B, served via vLLM, 300 problems): the full Level-3 design (\emph{full design}); a refactor that swaps the dataset-name router for an agnostic shape classifier (\emph{agnostic refactor}); an ablation with operators removed to bare multi-step CoT (\emph{operators ablated}); a variant adding deterministic stop and $K{=}3$ in-trajectory self-consistency on top (\emph{with stable termination}); and a code-routed extension that replaces the structured-state machinery with deterministic Python routing for AIME and FinQA (\emph{code-routed extension}).

\subsection{Heavyweight evaluation at 70B}
\label{sec:heavyweight-eval}

\begin{table}[h]
\caption{Heavyweight ReFlect on the 70B pair (300 problems each), with published-method baselines and the first lightweight breakout for context. \textbf{All rows use the same single-seed pilot-study scoring pipeline} for like-with-like comparison. Heavyweight variants: \emph{full design} = bugfixed Level-3 with structured state, 4 reflective operators, and regime FSM; \emph{agnostic refactor} = full design with the dataset-name router swapped for an agnostic shape classifier; \emph{operators ablated} = operators removed, bare multi-step CoT; \emph{with stable termination} = operators ablated, plus deterministic stop and $K{=}3$ in-trajectory self-consistency; \emph{code-routed extension} = structured-state machinery replaced with deterministic Python routing for AIME and FinQA.}
\label{tab:heavyweight-trajectory}
\centering
\small
\setlength{\tabcolsep}{6pt}
\begin{tabular}{@{}lccc@{}}
\toprule
\textbf{Method} & \textbf{Llama (\%)} & \textbf{Qwen (\%)} & \textbf{Pair-mean (\%)} \\
\midrule
\multicolumn{4}{@{}l}{\textit{Published baselines (single-seed, pilot scoring)}} \\
Direct CoT~\citep{wei2022chain} (Lvl 0)                                       & 17.8 & 12.2 & 15.0 \\
ReAct~\citep{yao2022react} (Lvl 1)                                            & 15.4 & 13.0 & 14.2 \\
Self-Refine~\citep{madaan2023selfrefine} (Lvl 2)                              & 14.3 & 12.4 & 13.4 \\
Reflexion~\citep{shinn2023reflexion} (Lvl 2)                                  & 20.9 & 17.5 & 19.2 \\
\midrule
\multicolumn{4}{@{}l}{\textit{Heavyweight ReFlect family progression (Figure~\ref{fig:rq3-ablation-summary} red bars)}} \\
\textsc{HW-Full} (full design)                                                 & 16.7 & 13.3 & 15.0 \\
\textsc{HW-Agnos} (agnostic refactor)                                          & 19.0 & 17.7 & 18.3 \\
\textsc{HW-Bare} (operators ablated)                                           & 20.0 & 13.5 & 16.7 \\
\textsc{HW-Best} (with stable termination)                                     & 20.2 & 17.2 & 18.7 \\
\textsc{HW-Code} (code-routed extension)                                       & 26.9 & 29.2 & 28.0 \\
\bottomrule
\end{tabular}
\end{table}

At the 70B scale, the heavyweight harness already matches or exceeds three of the four published baselines under like-with-like single-seed pilot scoring. Swapping the dataset-name router for an agnostic shape classifier (\emph{agnostic refactor}, 18.3\%) gives a small lift over the full design (15.0\%) but plateaus at the structured-state ceiling. The operator-ablation progression --- operators ablated 16.7\% $\to$ with stable termination 18.7\% --- climbs monotonically through the same ceiling, and the strongest variant (18.7\%) clears Direct CoT (15.0\%), ReAct (14.2\%), and Self-Refine (13.4\%) by 4.5 to 5.3~pp, sitting essentially level with Reflexion (19.2\%, within seed-variance noise). The final pilot-study iteration (\emph{code-routed extension}) then jumps to 28.0\% pair-mean (+9.3~pp over the structured-state plateau, +8.8~pp over Reflexion) by replacing the structured-state machinery with deterministic Python routing for AIME and FinQA, motivating the polished lightweight redesign reported in RQ2 (\S\ref{sec:lightweight-section}).

\paragraph{Three diagnoses and the base-capability prerequisite.}\label{sec:heavyweight-diagnoses-section} Execution-trace analysis (full evidence in Appendix~\ref{app:heavyweight-diagnoses}) yields three diagnoses, all flowing from a single \emph{base-capability prerequisite}: the state-extraction LLM call produces zero evidence items on $84\%$ of problems; operators fire on $\leq 5\%$ of steps; the regime FSM never reaches \textsc{Consolidate} ($0\%$ convergence). At 70B scale, the backbone cannot reliably populate the structured state the design depends on, so the structured-state machinery (operators, regime FSM, dependency cascade) is largely dormant; the 4--5~pp lift over Direct CoT comes from the multi-step deterministic-stop loop, not from the reflective primitives the heavyweight design was built around.

\section{RQ2: Lightweight Harness}
\label{sec:lightweight-section}

\subsection{Lightweight harness instantiation: shape routing and tool registry}
\label{sec:lightweight}

The lightweight instantiation realizes the harness in its most principled form, distilled to the three principles: a deterministic shape classifier feeding a small registry of specialized tools, with each invocation self-contained and mechanically checkable (Figure~\ref{fig:architecture}). By holding no state across LLM calls and routing entirely through pure-Python code, every dispatch decision is trivially auditable and every failure mode is addressable by editing a tool rather than revising a prompt. The classifier $\textsc{Shape}: x \mapsto s$ inspects problem-intrinsic features (unified-diff scaffolding, tabular context, action verbs, inference rules, code-shaped output, etc.; the classifier never references dataset names) and assigns one of seven computational shapes (Figure~\ref{fig:architecture}a); the registered tool then handles the problem end-to-end through a single \texttt{solve(problem, context, model)} interface (Figure~\ref{fig:architecture}b lists the seven tools and their mechanisms), with three tools enforcing format validators that reject malformed outputs and retry (the Validate Loop, Figure~\ref{fig:architecture}c), and \textsc{Fallback} catching anything the classifier cannot place. We evaluate two layered configurations of this design: \emph{Lightweight ReFlect (no domain tools)} uses only the dataset-agnostic core (the classifier plus the four generic per-shape tools in Figure~\ref{fig:architecture}b's top row), while \emph{Full ReFlect} (\S\ref{sec:main-results}) extends the registry with two task-specific tools (an ALFRED action-trace state tracker and a SWE-bench diff verifier; the bottom-left two cards in Figure~\ref{fig:architecture}b) to push the headline numbers.

\subsection{Main results}
\label{sec:main-results}
\label{sec:experiments}

\begin{table}[t]
\caption{Main results: task success rate (\%) across six domains (50 problems each) and six models spanning three capability tiers. SWE-bench uses a tiered structural-quality scorer (Appendix~\ref{app:swebench-scorer}); QASPER uses token-level F1; all other domains use exact/partial match. $\dagger$\,SWE-bench measures patch validity only, not bug-fix correctness, and is excluded from Avg. $\ddagger$\,Avg over the five non-SWE domains; the 6-domain version (21.3--29.3 Direct, 48.2--61.3 Full ReFlect; lifts $+19$ to $+39$~pp) is plotted in Figure~\ref{fig:capability-ladder}.}
\label{tab:main-results}
\centering
\small
\setlength{\tabcolsep}{4pt}
\begin{tabular}{@{}llcccccc|c@{}}
\toprule
\textbf{Tier} & \textbf{Model} & \textbf{SWE$^\dagger$} & \textbf{QASP} & \textbf{Proof} & \textbf{AIME} & \textbf{ALF} & \textbf{FinQA} & \textbf{Avg$^\ddagger$} \\
\midrule
\multicolumn{9}{l}{\textit{Direct CoT (no harness)}} \\
\midrule
Small & Claude Haiku 4.5   &  0.0 &  7.1 & 30.0 & 12.0 &  4.5 & 74.0 & 25.5 \\
Small & gpt-4o-mini        &  0.0 &  8.6 & 40.0 &  0.0 &  0.0 & 82.0 & 26.1 \\
70B   & Llama-3.3-70B      &  0.0 & 12.1 & 54.0 & 12.0 &  0.6 & 78.0 & 31.3 \\
70B   & Qwen2.5-72B       &  0.0 & 10.3 & 78.0 &  4.0 &  1.6 & 82.0 & 35.2 \\
Front.& GPT-4o             &  0.0 &  9.2 & 66.0 &  2.0 &  0.0 & 78.0 & 31.0 \\
Front.& Claude Sonnet 4.5  &  0.0 &  7.8 & 34.0 & 10.0 &  1.5 & 82.0 & 27.1 \\
\midrule
\multicolumn{9}{l}{\textit{Full ReFlect (shape-routed harness)}} \\
\midrule
Small & Claude Haiku 4.5   & 84.0 &  6.3 & 74.0 & 42.0 & 35.8 & 82.0 & 48.0 \\
Small & gpt-4o-mini        & 82.6 & 11.7 & 60.0 & 18.0 & 34.8 & 82.0 & 41.3 \\
70B   & Llama-3.3-70B      & 83.2 & 13.0 & 58.0 & 22.0 & 34.3 & 84.0 & 42.3 \\
70B   & Qwen2.5-72B       & 81.6 & 10.0 & 68.0 & 18.0 & 39.2 & 74.0 & 41.8 \\
Front.& GPT-4o             & 87.2 & 10.6 & 78.0 & 22.0 & 40.0 & 82.0 & 46.5 \\
Front.& Claude Sonnet 4.5  & 85.6 &  7.5 & 96.0 & 46.0 & 48.7 & 84.0 & 56.4 \\
\bottomrule
\end{tabular}
\end{table}

Full ReFlect achieves a 41 to 56\% correctness average across the five accuracy-scored domains, a $+7$ to $+29$~pp lift over Direct CoT on every model tested; SWE-bench additionally rises from 0\% to 82 to 87\% structural quality. Three findings stand out.

\emph{(1) Floor-domain tools deliver the largest per-domain gains.} ALFRED jumps from 0--5\% (Direct) to 34--49\% (Full ReFlect) via a pure-Python state tracker; SWE-bench moves from 0\% to 82--87\% via a diff verifier that forces syntactically valid patches. These two tools are task-specific (registered only for the ALFRED and SWE-bench shapes; the dataset-agnostic Lightweight ReFlect (no domain tools) variant excludes them) and compensate for capabilities the LLM lacks (physical precondition checking, diff-format compliance) regardless of model scale.

\emph{(2) The harness lift is largest on the weakest models — capability-compensation slope $=-1.69$.} Harness lift is inversely proportional to base model capability (slope $-1.69$~pp per pp, $r=-0.76$; Figure~\ref{fig:capability-ladder}d); each pp of capability loss is mechanically recovered by 1.69~pp of harness lift, so a small/cheap model paired with the harness closes most of the gap to a much larger model. Concretely, Haiku 4.5 ($25.5\% \to 48.0\%$, $+22.5$~pp) gains more than Qwen2.5-72B ($35.2\% \to 41.8\%$, $+6.6$~pp; panel~a). A near-uniform 19--22~pp of this lift comes from two task-specific deterministic-Python tools added on top of the dataset-agnostic shape-classifier backbone — the ALFRED state tracker ($+31$ to $+47$~pp) and the SWE-bench diff verifier ($+82$ to $+87$~pp structural) — both of which operate independently of model capability and add no LLM calls (panel~b). Re-fitting on the four LLM-driven domains (excluding ALFRED and SWE-bench) shifts the slope to $-1.66$ ($r=-0.84$, $p=0.036$; panel~c), so capability-compensation survives and tightens when deterministic-tool contribution is removed (per-model breakdown and an over-restrictive 3-domain refit are in Appendix~\ref{app:rq3-family-summary}).


\label{sec:systematic-failures-section}
\label{sec:capability-ladder}
\label{sec:comparison}

\section{RQ3: Ablation Experiments}
\label{sec:heavyweight-vs-lightweight}


\subsection{Token-normalized efficiency}
\label{sec:compute-matched}

On the 70B subset of the main grid (Llama-3.3-70B and Qwen2.5-72B, 300 problems each), the lightweight harness dominates every multi-call baseline on both accuracy and token cost (full numbers in Table~\ref{tab:compute-matched}, Appendix~\ref{app:serving}). Apples-to-apples on vLLM bf16, the code-routed backbone reaches 29.3\% pair-mean accuracy, beating Reflexion by $+1.4$~pp at $3.5\times$ lower tokens (\$2.79 vs \$10.26 per 100 correct), Self-Refine by $+2.1$~pp, and Direct CoT by $+2.4$~pp. The dataset-agnostic shape-routing extension then reduces per-problem tokens further while raising accuracy: \emph{Lightweight ReFlect (no domain tools)} runs at 2,939 tokens / 28.9\%, and Full ReFlect runs at 1,993 tokens / 48.8\% — a $+19.5$~pp gain over the code-routed backbone at $4.6\times$ fewer tokens, and a $+21.9$~pp gain over Direct CoT at near-identical token cost (\$0.36 vs \$0.66 per 100 correct), making Full ReFlect the cheapest method tested at the highest accuracy. 
\subsection{Cross-family ablation analysis}
\label{sec:rq3-family-summary}

\begin{figure}[t]
\centering
\includegraphics[width=0.85\linewidth]{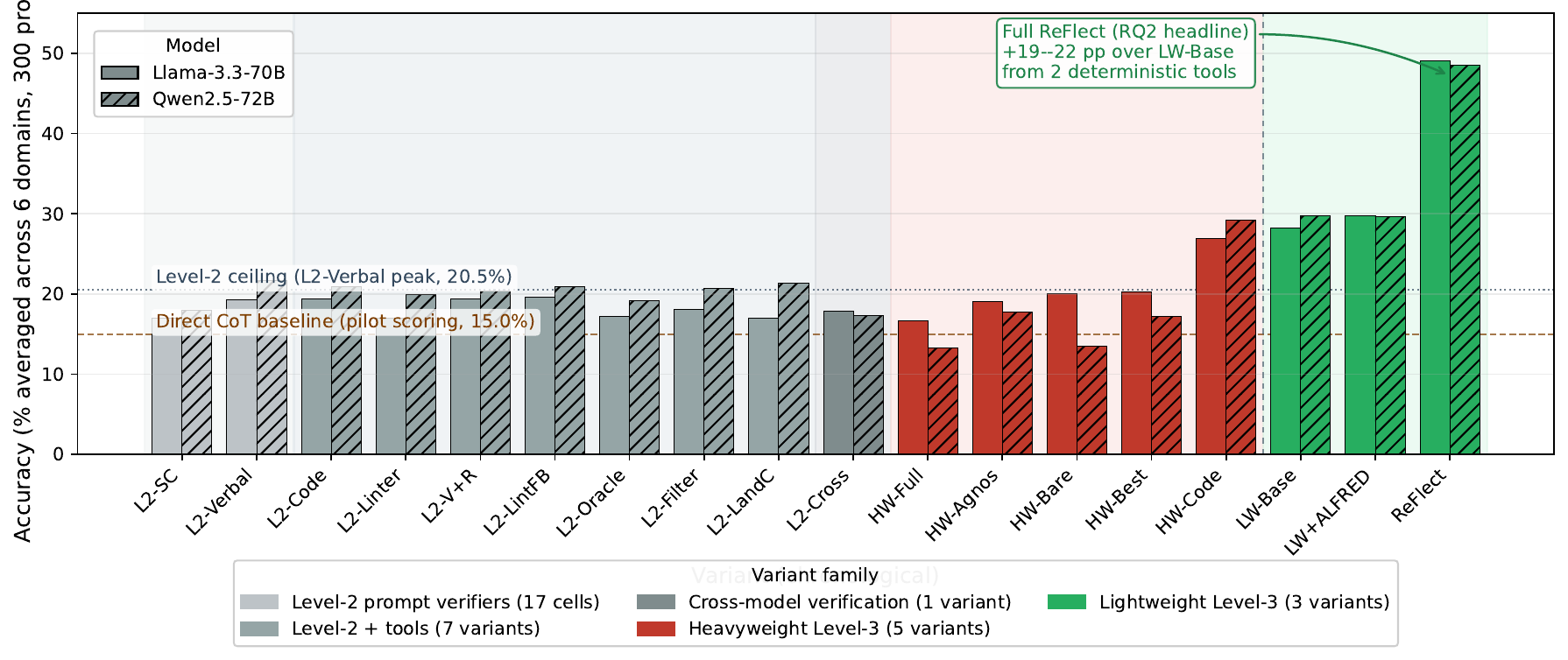}
\caption{28-variant RQ3 ablation on the 70B pair (Llama solid, Qwen hatched), extended with the polished lightweight progression on the right. Color: gray $=$ Level-2 prompt verifiers; \textbf{red $=$ Heavyweight ReFlect family} (5 variants in Table~\ref{tab:heavyweight-trajectory} order: \textsc{HW-Full}, \textsc{HW-Agnos}, \textsc{HW-Bare}, \textsc{HW-Best}, \textsc{HW-Code}; values match Table~\ref{tab:heavyweight-trajectory} under single-seed pilot scoring); green $=$ polished Lightweight ReFlect family (3 variants under RQ2 framework scoring). Vertical dashed gray separator marks the vLLM/API serving boundary; dotted line at 20.5\% marks the best Level-2 variant (L2-Verbal peak); dashed brown line at 15.0\% marks the Direct CoT baseline under pilot scoring (Table~\ref{tab:heavyweight-trajectory}). X-axis tag definitions are in Appendix~\ref{app:variant-sweep}, Table~\ref{tab:variant-tags}.}
\label{fig:rq3-ablation-summary}
\end{figure}

\emph{(a) The Level-2 ceiling is mechanism-invariant.} The three Level-2 families in Figure~\ref{fig:rq3-ablation-summary} (gray bars) cover all 24 prompt-level variants — verbal SC (L2-SC, L2-Verbal), tool-augmented (L2-Code, L2-Linter, L2-Oracle, L2-Filter, L2-L$\wedge$C), and cross-model (L2-Cross). All fall in the same narrow band (Llama 17.0--19.6\%, Qwen 17.3--21.7\%); no Level-2 variant exceeds the L2-Verbal peak of 20.5\%. Verifier FP rate stays in the 76--98\% band regardless of mechanism (Appendix~\ref{app:error-correction}); the L2-Cross row rules out correlated errors. The ceiling is a property of asking the model to verify its own free-text output, not of any particular verifier mechanism.

\emph{(b) Heavyweight progression and the code-routed jump.} The red bars (Heavyweight ReFlect family, Table~\ref{tab:heavyweight-trajectory} order under pilot scoring) trace the five-step progression: structured-state variants (\textsc{HW-Full}, \textsc{HW-Agnos}, \textsc{HW-Bare}, \textsc{HW-Best}) plateau at 15.0--18.7\% pair-mean, matching but not exceeding the Level-2 ceiling and Direct CoT (15\% line); the fifth bar (\textsc{HW-Code}, the code-routed extension) then jumps to 28.0\% pair-mean (26.9\% Llama / 29.2\% Qwen), a $+9.3$~pp gain over \textsc{HW-Best}. The lift is mechanically attributable to deterministic Python execution on AIME and FinQA --- the only mechanism that escapes the LLM-judged-verification ceiling because it does not ask the model to verify anything --- and it is the bridge to the polished lightweight redesign reported in RQ2 (the green bars, on the right of the serving boundary).

Taken together, these findings sharply constrain what carries the gain at 70B: prompt-level Level-2 verifiers all hit the same 20.5\% / 76\%-FP ceiling regardless of mechanism; the structured-state arm of the heavyweight family (\textsc{HW-Full}, \textsc{HW-Agnos}, \textsc{HW-Bare}, \textsc{HW-Best}; Figure~\ref{fig:rq3-ablation-summary}) plateaus at 15.0--18.7\% under pilot scoring (Table~\ref{tab:heavyweight-trajectory}), matching ordinary baselines but never exceeding the Level-2 ceiling; adding more reasoning calls without changing computational regime (Self-Refine, Reflexion) saturates at the same Level-2 plateau. Only \textsc{HW-Code}, which abandons the structured-state machinery for deterministic Python routing on AIME and FinQA, breaks through to 28.0\% pair-mean --- mechanically because Python execution is the only verification mechanism that does not ask the model to verify its own output.
\section{Discussion and limitations}
\label{sec:discussion}

ReFlect wraps a backbone LLM and converts open-ended failure into structurally-bounded computation (Figure~\ref{fig:architecture}). At 70B scale, the heavyweight structured-state arm (\textsc{HW-Full}, \textsc{HW-Agnos}, \textsc{HW-Bare}, \textsc{HW-Best}; Table~\ref{tab:heavyweight-trajectory}) matches but does not exceed ordinary baselines: operators fire on $<$5\% of steps because state-extraction yields zero evidence on 84\% of problems, so the modest 4--5~pp lift over Direct CoT comes from the multi-step deterministic-stop loop rather than the reflective primitives. The code-routed extension (\textsc{HW-Code}) jumps to 28.0\% pair-mean by abandoning that machinery for deterministic Python routing on AIME and FinQA. The lightweight design (\S\ref{sec:lightweight-section}; Figure~\ref{fig:architecture}) clears the structured-state prerequisite entirely and delivers 41--56\% accuracy across six models at \$0.36 per 100 correct, dominating Direct CoT on both accuracy and cost (Table~\ref{tab:compute-matched}). ReFlect composes naturally with workspace-reconstruction substrates of the IterResearch~\citep{chen2026iterresearch} or CoALA~\citep{sumers2024cognitive} kind (Appendix~\ref{app:discussion-extra}).

The present results delineate three concrete avenues for follow-on study. First, the heavyweight design exposes a measurable \emph{base-capability frontier}: at 70B scale the structured state cannot be populated reliably, which yields a falsifiable prediction that frontier-scale backbones (Claude Sonnet 4.5, GPT-4o) should clear the prerequisite — a hypothesis our framework makes directly testable (\S\ref{sec:heavyweight-diagnoses-section}). Second, the dataset-agnostic router pairs with task-shaped specialist tools, surfacing a transparent extension surface: when a domain is bottlenecked by an LLM capability the harness does not yet target (e.g., QASPER span extraction at 6--13\%), accuracy is bounded by that capability rather than by the harness, and adding a new shape and tool is a drop-in operation.

\section{Conclusion and Future Work}
\label{sec:conclusion}

We introduced ReFlect, a \emph{harness} system for LLM reasoning: reliability comes from the wrapper, not the prompt. A 360-run motivating pilot (\S\ref{sec:motivation}) and a 28-variant ablation (\S\ref{sec:heavyweight-vs-lightweight}, Figure~\ref{fig:rq3-ablation-summary}) establish that prompt-level self-critique fails systematically at 70B scale: $\geq$90\% boilerplate reflections, $\leq$1.7\% course correction, and verifier false-positive rates in the 76--98\% band regardless of mechanism. The heavyweight family progression (Table~\ref{tab:heavyweight-trajectory}: \textsc{HW-Full} $\to$ \textsc{HW-Agnos} $\to$ \textsc{HW-Bare} $\to$ \textsc{HW-Best}) plateaus at 15.0--18.7\% under structured-state machinery, while the in-family code-routed extension (\textsc{HW-Code}) jumps to 28.0\% by abandoning that machinery for deterministic Python routing --- the bridge to RQ2's polished lightweight redesign (Figure~\ref{fig:architecture}), which delivers 41--56\% accuracy across six models. ReAct made reasoning interactive; Self-Refine and Reflexion made it self-critiquing; ReFlect makes it \emph{harnessable} --- not by asking the model to think more, but by routing each failure into the deterministic machinery that can resolve it.

Three directions follow from these results. First, the base-capability prerequisite (\S\ref{sec:heavyweight-diagnoses-section}) makes a falsifiable scaling prediction: evaluating the heavyweight design with frontier backbones (Claude Sonnet 4.5, GPT-4o) tests whether structured-state operators become viable once state extraction is no longer the bottleneck. Second, learning the shape-routing decision online — via a trained classifier or a meta-controller over the tool registry — would extend ReFlect from a hand-built dispatcher to a self-extending harness while preserving the determinism that drives its reliability. Third, the harness layer composes with workspace-reconstruction substrates (IterResearch, CoALA), suggesting a meta-harness in which ReFlect handles deterministic detection-and-recovery while a workspace substrate handles long-context memory.

\section*{Acknowledgments}

We thank the maintainers and contributors of the open-source benchmarks that made this work possible: SWE-bench Lite~\citep{jimenez2024swebench}, QASPER~\citep{dasigi2021qasper}, ProofWriter~\citep{tafjord2021proofwriter}, AIME~\citep{aime2024}, ALFRED~\citep{shridhar2020alfred}, and FinQA~\citep{chen2021finqa}. We also thank the open-weights model teams whose models served as the 70B backbones for our pilot and ablation experiments: Llama-3.3-70B-Instruct~\citep{llama2024llama33} and Qwen2.5-72B-Instruct~\citep{qwen2024qwen25}. The inference infrastructure for the 70B vLLM-served evaluations relied on vLLM~\citep{kwon2023vllm}; the 6-model capability-ladder evaluations were served via the public Together.ai, OpenRouter, OpenAI, and Anthropic APIs.

Funding-source information will be released in the camera-ready version.

\paragraph{Use of AI tools.}
Large language models were used \emph{only} for grammar and style checking of the manuscript text. They were not used for ideation, experimental design, code generation (for the harness implementation, the analysis pipelines, or the figure-rendering notebooks), results analysis, or content authorship. All scientific claims, code, and data analyses are the authors' own.

\bibliographystyle{plainnat}
\bibliography{references}
\appendix

\section{Related work: extended per-paradigm discussion}
\label{app:related-detail}

The condensed §\ref{sec:related} omits per-method detail and the worked IterResearch comparison foreshadowed in §\ref{sec:taxonomy}. Both are restored here.

\paragraph{Single-pass generation.}
Chain-of-thought prompting~\citep{wei2022chain} and its inference-time variants (self-consistency~\citep{wang2023selfconsistency}, least-to-most~\citep{zhou2023leasttomost}) produce reasoning traces in a single forward pass. The training-time variant STaR~\citep{zelikman2022star} bootstraps CoT ability by fine-tuning on self-generated traces but shares the same inference-time limitation. These methods improve step-level quality but provide no mechanism for error detection or mid-task correction. Errors accumulate silently, and the system has no representation of what it believes, assumes, or has decided.

\paragraph{Interactive observation.}
ReAct~\citep{yao2022react} interleaves reasoning with environment actions, grounding intermediate steps in observations. Reasoning becomes interactive rather than purely generative. However, ReAct treats the token trajectory as implicit memory, offers no state compression, and cannot distinguish validated conclusions from unsupported assumptions. Tree of Thoughts~\citep{yao2024tree} adds search over reasoning paths but still operates on text-level representations without explicit state management.

\paragraph{LLM-judged self-correction.}
Self-Refine~\citep{madaan2023selfrefine} separates generation from critique via distinct LLM calls, iterating until convergence. This achieves a crucial decoupling (the critic evaluates the output holistically) but operates on complete output text, not structured state. It cannot track assumption dependencies, roll back to checkpoints, or intervene mid-task. Reflexion~\citep{shinn2023reflexion} adds cross-episode memory: the agent attempts a full episode, reflects on failure using environment feedback, and retries from scratch with that context. This is powerful when binary feedback is available (code tests, game scores) but inapplicable to open-ended tasks lacking such signals, and discards all within-episode progress on retry. CRITIC~\citep{gou2024critic} leverages external tools for verification but still issues critique as text. Our pilot's ``Minimal ReFlect'' (inline mid-task self-critique in the same generation stream) is the weakest member of this family; its systematic failure motivates the externalization argued for in \S\ref{sec:taxonomy}.

\paragraph{State management for long-horizon agents.}
IterResearch~\citep{chen2026iterresearch} addresses context suffocation through iterative workspace reconstruction, maintaining a bounded state $(q, M_t, \{a_{t-1}, \mathit{TR}_{t-1}\})$ where $M_t$ is a free-text report synthesized by the model at each round. This achieves O(1) state size and enables interaction scaling to 2{,}048 turns with consistent performance gains. However, the report is an unstructured text synthesis: a \emph{memory}, not a \emph{harness}. The system has no programmatic contradiction detection, no structural validators, no deterministic dispatch. The distinction is analogous to having \emph{good notes} versus having a \emph{quality-control process}: good notes prevent losing important information; quality control prevents building on flawed reasoning.

\paragraph{Worked example: where IterResearch and a structural harness differ.}
Suppose in round 5, an agent assumes ``revenue grew 15\% YoY'' based on a misread table, and this assumption gets baked into the report. In rounds 6 to 10, all analysis builds on it. Under IterResearch, the flawed assumption is embedded in the report's natural-language text: the model \emph{might} notice contradictory evidence later and revise the report, but this requires it to (a) detect the contradiction, (b) trace it to the original assumption, and (c) update all dependent conclusions. Our pilot shows 70B models do this $\leq$2\% of the time. Under ReFlect-heavyweight, the assumption is tracked as $a_5$ with \texttt{status=active} and linked dependents $[e_7, e_8, d_3]$; when contradictory evidence arrives, the conflict is detected programmatically, \textsc{Inspect} diagnoses it as critical, and \textsc{Transform} retracts $a_5$ with cascade. (At 70B, the heavyweight design fails to deliver this benefit because 70B models cannot reliably populate the state object that the cascade requires; see \S\ref{sec:heavyweight-vs-lightweight} and Appendix~\ref{app:heavyweight-diagnoses}.)

\section{Pilot study: per-domain accuracy table}
\label{app:pilot-results-table}

\begin{table}[h]
\caption{Pilot study results. Q = Qwen2.5-72B, L = Llama-3.3-70B. Minimal ReFlect never wins a single category; it ties or loses on every domain for both models. Bold indicates the best result per metric and model.}
\label{tab:pilot-results}
\centering
\small
\begin{tabular}{@{}llcccccc@{}}
\toprule
& & \multicolumn{2}{c}{\textbf{Direct}} & \multicolumn{2}{c}{\textbf{ReAct}} & \multicolumn{2}{c}{\textbf{Min.\ ReFlect}} \\
\cmidrule(lr){3-4} \cmidrule(lr){5-6} \cmidrule(lr){7-8}
\textbf{Domain} & \textbf{Metric} & Q & L & Q & L & Q & L \\
\midrule
AIME & Exact match & \textbf{10} & \textbf{20} & \textbf{10} & 10 & 0 & 0 \\
ProofWriter & Answer match & 30 & 40 & 30 & \textbf{50} & 30 & 40 \\
ALFRED & Action recall & \textbf{100} & \textbf{92} & 90 & 90 & 93 & 88 \\
ALFRED & Order correct & \textbf{80} & \textbf{70} & 60 & 50 & 60 & 60 \\
FinQA & Exact (1\%) & \textbf{10} & \textbf{10} & 0 & 0 & 0 & 0 \\
FinQA & Close (5\%) & \textbf{10} & \textbf{10} & 0 & 10 & 0 & 0 \\
\bottomrule
\end{tabular}
\end{table}

\section{Pilot study: root-cause analysis (full detail)}
\label{app:pilot-rootcauses}

The five root causes from \S\ref{sec:root-causes}, expanded:

\paragraph{1. Repetition loops (catastrophic).}
Models enter degenerate loops repeating the same phrase dozens to hundreds of times, consuming the entire output budget. Prevalence: 7\% of Qwen ReFlect runs, 23\% of Llama ReFlect runs, near-zero for Direct/ReAct. Causal chain: reflection prompt $\to$ model reaches answer $\to$ self-query of the form ``is this correct?'' or ``verify again'' $\to$ re-states answer $\to$ infinite loop $\to$ truncation.

\paragraph{2. Boilerplate reflections (systemic).}
Over 90\% of reflection blocks are formulaic affirmations providing zero corrective signal. A representative block recurs nearly verbatim across episodes:
\begin{quote}\small\ttfamily
Reflection:\\
\hphantom{xx}- State check: I have computed the intermediate result.\\
\hphantom{xx}- Consistency check: No contradictions with earlier steps.\\
\hphantom{xx}- Assumption check: No unsupported assumptions.\\
\hphantom{xx}- Direction check: On track toward the goal.\\
\hphantom{xx}- Decision: Continue.
\end{quote}
In ALFRED and FinQA, 100\% of Qwen's reflection blocks are boilerplate. Llama is worse: 85\% of runs contain zero proper reflection blocks.

\paragraph{3. Zero course correction (systemic).}
Reflection almost never leads to a changed approach: 2\% correction rate for Qwen (1/60), 0\% for Llama (0/60). In the single Qwen case where reflection changed the answer, the change was from one wrong answer to a different wrong answer.

\paragraph{Statistical caveat.}
The pilot uses 60 problems per (model, method) pair (10 per domain $\times$ 6 domains, single seed). Wilson 95\% confidence intervals on the headline rates: course correction $\leq 2\%$ corresponds to a CI of $[0\%, 6.0\%]$ (Llama: 0/60) and $[0.3\%, 8.9\%]$ (Qwen: 1/60); $90\%+$ boilerplate (90/100 blocks) is $[82.6\%, 94.5\%]$; the 7 to 23\% repetition-loop range covers $[2.6\%, 15.9\%]$ (Qwen 4/60) to $[14.4\%, 35.4\%]$ (Llama 14/60). The CIs are wide for the smallest-count claims, but the qualitative finding (reflection rarely produces course correction or substantive content) holds across the credible range.

\paragraph{4. Token budget starvation.}
Reflection's structural overhead (+30--60\% more generated text) pushes outputs toward the generation limit. Llama ReFlect was truncated on 25\% of runs.

\paragraph{5. Format non-compliance.}
Llama-3.3-70B largely ignores the structured reflection format, substituting its own inline checklists that are identical boilerplate after every observation.

\section{Pilot study: detailed structural metrics}
\label{app:pilot-structural}

Table~\ref{tab:pilot-structural} reports structural metrics from the 360-run pilot study. Despite worse accuracy, Minimal ReFlect shows dramatically different trajectory structure: more reflections, more backtracking, and substantially more contradiction detection.

\begin{table}[h]
\caption{Structural metrics from the pilot study (averaged across domains). Q = Qwen2.5-72B, L = Llama-3.3-70B.}
\label{tab:pilot-structural}
\centering
\small
\begin{tabular}{@{}lcccccc@{}}
\toprule
& \multicolumn{2}{c}{\textbf{Direct}} & \multicolumn{2}{c}{\textbf{ReAct}} & \multicolumn{2}{c}{\textbf{Min.\ ReFlect}} \\
\cmidrule(lr){2-3} \cmidrule(lr){4-5} \cmidrule(lr){6-7}
\textbf{Metric} & Q & L & Q & L & Q & L \\
\midrule
Avg.\ thoughts & 0.0 & 0.0 & 4.3 & 3.4 & 4.7 & 5.3 \\
Avg.\ actions & 0.6 & 0.6 & 6.5 & 4.7 & 5.9 & 6.8 \\
Avg.\ reflections & 0.0 & 0.0 & 0.0 & 0.0 & 1.7 & 2.5 \\
Has backtrack (\%) & 2 & 3 & 3 & 5 & 10 & 13 \\
Has contradiction (\%) & 8 & 13 & 10 & 12 & 52 & 42 \\
Has final answer (\%) & 55 & 8 & 98 & 95 & 100 & 98 \\
\bottomrule
\end{tabular}
\end{table}

\section{Reflection quality audit}
\label{app:reflection-audit}

Table~\ref{tab:reflection-audit} audits the quality of reflection blocks produced by Qwen2.5-72B under the Minimal ReFlect method. Over 90\% of reflection blocks are formulaic boilerplate providing zero corrective signal.

\begin{table}[h]
\caption{Reflection block audit for Qwen2.5-72B Minimal ReFlect. Substantive = reflection identifies a genuine issue or leads to a course change. Boilerplate = formulaic ``no issues / on track / continue.''}
\label{tab:reflection-audit}
\centering
\small
\begin{tabular}{@{}lccc@{}}
\toprule
\textbf{Domain} & \textbf{Total blocks} & \textbf{Substantive (\%)} & \textbf{Boilerplate (\%)} \\
\midrule
AIME & 20 & 25 & 70 \\
ALFRED & 19 & 0 & 100 \\
FinQA & 10 & 0 & 100 \\
ProofWriter & 20 & 5 & 95 \\
QASPER & 10 & 0 & 100 \\
SWE-bench & 21 & 10 & 86 \\
\midrule
\textbf{Overall} & 100 & 8 & 90 \\
\bottomrule
\end{tabular}
\end{table}

\section{ReFlect framework: design principles}
\label{app:principles}

The two ReFlect instantiations evaluated as RQ1 (heavyweight, \S\ref{sec:method}) and RQ2 (lightweight, \S\ref{sec:lightweight-section}) follow the same three design principles. We expand each here.

\paragraph{Principle 1: Harness over scaffold.}
The system's reliability comes from the wrapper around the model, not from a more sophisticated prompt. The harness owns problem classification, tool dispatch, format validation, and retry policy as deterministic code; the model is invoked only inside structurally-constrained slots whose outputs the harness can mechanically check. This separates \emph{what the model is good at} (generating candidates inside a constrained shape) from \emph{what the model is bad at} (self-monitoring, format compliance, deterministic verification), and gives the latter to non-LLM machinery.

\paragraph{Principle 2: Structural routing over LLM-judged self-correction.}
The signal that does the work in our framework is not the model's own self-critique (which our pilot shows degenerates into boilerplate at 70B; \S\ref{sec:motivation}), but a deterministic routing decision the harness makes \emph{outside} the model. The harness inspects problem-intrinsic features (presence of tables, action verbs, inference rules, code-shaped output requirements) and assigns each problem to a computational regime. Diverse failure modes collapse onto a small finite set of regime mismatches, and each regime mismatch maps to a deterministic intervention (retry under stricter format, fall back to self-consistency, escalate to a more general tool). The model is never asked to evaluate its own output in free-text; it is asked only to generate inside a structurally-bounded slot whose validity the harness can mechanically check.

\paragraph{Principle 3: Computational regime over reasoning style.}
Heterogeneous reasoning tasks are not solvable by a single computational machinery. The harness shifts problems between regimes (deterministic-symbolic via SymPy, tabular-arithmetic via pandas/eval, logical-inference via forward chaining, evidence-extraction via retrieval, procedural-validation via precondition checker, artifact-generation via format validator), and within a regime, falls back to self-consistency when validation fails. What changes between problems is not the model's \emph{thinking style} but the \emph{machinery the model is plugged into}. The 70B LLM-judged verification ceiling (76 to 98\% verifier FP rate, 14 to 21\% accuracy across 24 Level-2 variants) is a property of asking the model to verify in free-text; the ceiling lifts only when problems are routed \emph{out} of the LLM-judged regime entirely, into a deterministic computation whose result the harness can trust.

\section{Heavyweight harness instantiation: full design}
\label{app:heavyweight-detail}

This appendix expands \S\ref{sec:heavyweight} with the full schema, operator definitions, controller policy, regime-transition rules, and pseudocode for the heavyweight ReFlect instantiation; the empirical evaluation of this design (the Heavyweight ReFlect family) is in \S\ref{sec:heavyweight-vs-lightweight} and Appendix~\ref{app:variant-sweep}. (Figure~\ref{fig:architecture} in the main body shows the lightweight architecture; the heavyweight design is given by Algorithm~\ref{alg:reflect} below.)

\paragraph{Reasoning state.}
The reasoning state $\mathcal{S}$ is a structured, mutable data object maintained outside the LLM:

\begin{equation}
\mathcal{S} = \bigl(\mathcal{G},\; \mathcal{A},\; \mathcal{E},\; \mathcal{D},\; \mathcal{C},\; \mathcal{T},\; \mathcal{K},\; r,\; u\bigr)
\end{equation}

\noindent where $\mathcal{G}$ is a hierarchical goal tree (each goal has status, parent/child links), $\mathcal{A}$ is a set of tracked assumptions (each with justification, status $\in \{\text{active}, \text{validated}, \text{retracted}\}$, and dependency links to other elements), $\mathcal{E}$ is sourced evidence (with provenance and confidence levels), $\mathcal{D}$ is strategic decisions (with rationale and reversibility flags), $\mathcal{C}$ is detected conflicts between state elements, $\mathcal{T}$ is a compressed trajectory (recent steps verbatim, older steps summarized), $\mathcal{K}$ is a set of checkpoints for rollback, $r \in \{\textsc{Explore}, \textsc{Execute}, \textsc{Verify}, \textsc{Recover}, \textsc{Consolidate}\}$ is the current regime, and $u \in [0,1]$ is a composite uncertainty estimate.

\paragraph{Assumption tracking.}
The key innovation in the state design is treating assumptions as first-class objects with dependency links. When assumption $a_i$ is retracted, all elements that depend on $a_i$ (evidence derived from it, decisions based on it, sub-goals predicated on it) are automatically flagged or retracted in cascade. This addresses the pilot's observation that models silently adopt unsupported assumptions that pollute all downstream reasoning.

\paragraph{Uncertainty estimation.}
The composite uncertainty $u$ is derived from four normalized signals: (1) ratio of unvalidated assumptions, (2) density of unresolved conflicts (saturating at 3), (3) ratio of low-confidence evidence, and (4) ratio of blocked goals. The controller's uncertainty threshold (default $\theta_u = 0.6$) determines when \textsc{Inspect} is triggered.

\paragraph{State extraction.}
After each reasoning step, a lightweight \emph{separate} LLM call extracts structured elements (evidence, assumptions, decisions, goal updates, conflicts) from the free-text output using a domain-agnostic extraction prompt. The base reasoner thinks freely in natural language; structure is imposed after the fact, preserving reasoning quality.

\paragraph{Compiled view.}
The function $\texttt{compile\_view}(\mathcal{S}, r)$ constructs a regime-shaped prompt from the current state. In \textsc{Execute} mode, completed goals and resolved conflicts are hidden to focus forward progress. In \textsc{Verify} mode, assumptions are framed as ``claims to challenge,'' priming adversarial checking. In \textsc{Recover} mode, the diagnosis and failed goals are shown prominently. The base LLM's behavior changes because its input changes, not because of a special instruction.

\paragraph{Reflective operators.}
Four operators form a minimal action space for the heavyweight harness. Each is a separate programmatic intervention, not an inline prompt: \textsc{Inspect} abstracts raw failures into structured diagnostics; the structured state serves as a persistent recovery representation; \textsc{Transform} executes targeted interventions based on these representations; and \textsc{Stabilize} consolidates successful recoveries into the state.

\paragraph{\textsc{Inspect}: diagnose state quality.}
A separate LLM call receives only the state object (not the full trajectory) and checks for: (1) unsupported assumptions, (2) contradictions between state elements, (3) stalled progress, and (4) unjustified confidence. It returns a \emph{structured diagnostic} (not free-form critique) with fields: failure type $\in \{\text{logic}, \text{arithmetic}, \text{unsupported}, \text{incomplete}, \text{contradiction}, \text{stalled}\}$, affected state elements, severity, and overall health (\texttt{good} / \texttt{caution} / \texttt{critical}). This is the harness principle in action: diverse failure modes are actively compressed into a finite set of categories that map directly to operator interventions. This works where the pilot failed because the auditor examines a structured state object and produces a structured signal, not a 3000-token free-text response that degenerates into ``no issues found, continue.''

\paragraph{\textsc{Stabilize}: compress and consolidate.}
Mostly programmatic: (1) compress the trajectory (keep recent 5 steps verbatim, summarize older ones), (2) promote assumptions validated by evidence, (3) archive completed sub-goals, (4) create a checkpoint for potential rollback, (5) reset step counters. Keeps the working state lean and prevents context overflow.

\paragraph{\textsc{Transform}: intervene and correct.}
Modifies the state in response to diagnosed problems. For unsupported assumptions: retract and cascade to dependents. For contradictions: present both sides to the LLM for resolution, then apply the resolution programmatically. For stalled progress: rollback to the last checkpoint and replan with a fresh strategy. For overconfidence: downgrade confidence on flagged elements. This is where genuine course correction happens: the next reasoning step sees a \emph{genuinely different} state, not a narrative suggestion to ``reconsider.'' Crucially, the \textsc{Inspect} diagnosis and the modified state together form a \emph{reusable recovery state}: the system retains not just what went wrong (failure abstraction) but how it was fixed (recovery action), enabling better attribution and faster recovery when similar failures recur later in the trajectory.

\paragraph{\textsc{Diversify}: explore alternatives.}
When the best path forward is uncertain, fork the state into $N$ branches (default 3), run each forward for $K$ steps (default 5), then compare and select the most promising branch. An at-most-once policy and token budget gating prevent this expensive operator from dominating compute. If the budget is insufficient, the controller falls back to \textsc{Transform} with rollback.

\paragraph{Controller and regime switching.}
\label{sec:controller}
The controller is a rule-based scheduler that decides which operator to invoke after each reasoning step, based on priority-ordered trigger conditions:

\begin{enumerate}[leftmargin=*,itemsep=1pt]
\item \textbf{Critical:} New conflict detected $\to$ \textsc{Inspect}
\item \textbf{High:} Uncertainty $u > \theta_u$ $\to$ \textsc{Inspect}
\item \textbf{Medium:} Steps since last reflection $> K_{\max}$ $\to$ \textsc{Stabilize}
\item \textbf{Medium:} Steps since last progress $> K_{\text{stall}}$ $\to$ \textsc{Diversify}
\item \textbf{Low:} Pending major decision $\to$ \textsc{Inspect}
\end{enumerate}

\noindent If \textsc{Inspect} returns a \texttt{critical} diagnosis, \textsc{Transform} is invoked immediately. At most one operator fires per step; complex intervention chains emerge from sequential steps.

\paragraph{Regime transitions.}
The controller manages a five-state machine:
\begin{itemize}[leftmargin=*,itemsep=1pt]
\item $\textsc{Explore} \to \textsc{Execute}$: when a committed plan exists (strategy decision + active sub-goals).
\item $\textsc{Execute} \to \textsc{Verify}$: when $\geq$75\% of leaf goals are done with none blocked.
\item $\textsc{Execute} \to \textsc{Recover}$: when critical issues are detected.
\item $\textsc{Verify} \to \textsc{Consolidate}$: when the last inspection finds no issues.
\item $\textsc{Verify} \to \textsc{Recover}$: when verification fails.
\item $\textsc{Recover} \to \textsc{Execute}$: when uncertainty drops below 0.4 with no critical conflicts.
\end{itemize}

\paragraph{Main loop.}
Algorithm~\ref{alg:reflect} presents the heavyweight execution flow.

\begin{algorithm}[h]
\caption{\textsc{ReFlect-Heavyweight}$(x, \mathcal{M}, \textit{config})$}
\label{alg:reflect}
\begin{algorithmic}[1]
\Require Problem $x$, backbone LLM $\mathcal{M}$, controller config
\State $\mathcal{S} \gets \textsc{InitState}(x)$ \Comment{Parse problem into initial goal tree}
\State $\textit{ctrl} \gets \textsc{Controller}(\textit{config})$
\For{$t = 1, \dots, T_{\max}$}
    \State $\textit{prompt} \gets \texttt{compile\_view}(\mathcal{S},\; \textit{ctrl.regime})$ \Comment{Regime-shaped view}
    \State $y_t \gets \mathcal{M}(\textit{prompt})$ \Comment{Generate reasoning step}
    \State $\Delta_t \gets \textsc{Extract}(\mathcal{S}, y_t, \mathcal{M})$ \Comment{Separate extraction call}
    \State $\mathcal{S} \gets \mathcal{S} \oplus \Delta_t$ \Comment{Update state with extracted elements}
    \State $\textit{op} \gets \textit{ctrl}.\textsc{Step}(\mathcal{S})$ \Comment{Controller decides intervention}
    \If{$\textit{op} = \textsc{Inspect}$}
        \State $\textit{dx} \gets \textsc{Inspect}(\mathcal{S}, \mathcal{M})$
        \If{$\textit{dx}.\textit{health} = \texttt{critical}$}
            \State $\mathcal{S} \gets \textsc{Transform}(\mathcal{S}, \textit{dx}, \mathcal{M})$
        \EndIf
    \ElsIf{$\textit{op} = \textsc{Stabilize}$}
        \State $\mathcal{S} \gets \textsc{Stabilize}(\mathcal{S}, \mathcal{M})$
    \ElsIf{$\textit{op} = \textsc{Diversify}$}
        \State $\mathcal{S} \gets \textsc{Diversify}(\mathcal{S}, N, K, \mathcal{M})$
    \EndIf
    \State $\textit{ctrl}.\textsc{UpdateRegime}(\mathcal{S})$
    \If{$\mathcal{S}.\texttt{is\_complete}()$} \textbf{break} \EndIf
\EndFor
\State \Return $\mathcal{S}.\texttt{compile\_answer}()$
\end{algorithmic}
\end{algorithm}

\subsection{Lightweight algorithm}
\label{app:lightweight-algo}
For comparison, Algorithm~\ref{alg:lightweight} gives the lightweight harness pseudocode (Full ReFlect, \S\ref{sec:lightweight}).

\begin{algorithm}[h]
\caption{\textsc{ReFlect-Lightweight}$(x, \mathcal{M})$}
\label{alg:lightweight}
\begin{algorithmic}[1]
\Require Problem $x$, backbone LLM $\mathcal{M}$, tool registry $\mathcal{R}$
\State $s \gets \textsc{Shape}(x)$ \Comment{deterministic, problem-intrinsic}
\State \textit{tool} $\gets \mathcal{R}[s]$
\State $a \gets \textit{tool}.\textsc{Solve}(x, \mathcal{M})$ \Comment{may invoke validators and retries}
\If{$a = \texttt{None}$}
    \State $a \gets \mathcal{R}[\textsc{Fallback}].\textsc{Solve}(x, \mathcal{M})$
\EndIf
\State \Return $a$
\end{algorithmic}
\end{algorithm}

\section{Heavyweight underperformance: full diagnostic detail}
\label{app:heavyweight-diagnoses}

The three diagnoses summarized in \S\ref{sec:heavyweight-vs-lightweight}, drawn from execution-trace analysis on the full-design Heavyweight ReFlect runs:

\paragraph{Extract step is unreliable.}
The state-extraction LLM call produces zero evidence items on 84\% of problems (Llama: 92\%; Qwen: 76\%). 70B models' free-text reasoning rarely contains cleanly-structured evidence statements that the extractor can parse, leaving the state machinery acting on stubs. A fallback handles empty extraction, but downstream operators that depend on populated state objects then fire at $<$5\% rate.

\paragraph{Operators rarely fire.}
Even after extraction succeeds, operator firing rates remain low: \textsc{Inspect} fires on 5\% of steps, \textsc{Transform} on 5\%, \textsc{Diversify} on 0\%. The controller's trigger conditions (uncertainty $u > \theta_u$, conflict count, stalled-progress detector) depend on populated state objects that the model rarely produces. The four-operator action space collapses to a near-no-op scheduler.

\paragraph{Regime FSM is inert.}
No problem in the full-design Heavyweight ReFlect runs reaches \textsc{Consolidate}; convergence is 0\%. The five-state machine collapses to \textsc{Explore} $\leftrightarrow$ \textsc{Execute} for almost all problems. The transitions \textsc{Verify} $\to$ \textsc{Consolidate} (requires last inspection finds no issues) and \textsc{Recover} $\to$ \textsc{Execute} (requires uncertainty drops below 0.4 with no critical conflicts) never fire because the predicates that gate them are never satisfied, a direct consequence of the empty-state issue above.

\paragraph{Common cause.}
All three diagnoses share a single root: 70B models cannot reliably populate or reason about structured state objects. The heavyweight design encodes a base-capability prerequisite that this scale of model does not meet. Whether the prerequisite is met at higher capability remains open and is not addressed by our experiments.

\section{Detailed paradigm comparison}
\label{app:paradigm-comparison}

This appendix expands the per-paradigm structural-primitive comparison foreshadowed in \S\ref{sec:taxonomy} and \S\ref{sec:comparison}. Table~\ref{tab:comparison} compares ReFlect against six existing inference-time reasoning paradigms — CoT~\citep{wei2022chain}, ReAct~\citep{yao2022react}, Tree of Thoughts~\citep{yao2024tree}, Self-Refine~\citep{madaan2023selfrefine}, Reflexion~\citep{shinn2023reflexion}, and IterResearch~\citep{chen2026iterresearch} — across ten structural primitives. Cells annotated \cmark\textsuperscript{H} are present only in the heavyweight ReFlect instantiation (\S\ref{sec:heavyweight}); \cmark\textsuperscript{L} only in the lightweight instantiation (\S\ref{sec:lightweight}, the system actually evaluated as Full ReFlect); unmarked \cmark in both. Two takeaways. \textbf{(1) Structured-state primitives are unique to ReFlect's heavyweight instantiation among prior systems.} Assumption tracking with dependency cascade, rollback to checkpoints, branching/search over reasoning paths, and regime switching are absent from every other Level-0 to Level-2 paradigm; ReFlect contributes them as a coherent Level-3 design (which the paper then reports as a deliberate negative result at 70B, \S\ref{sec:method}). \textbf{(2) Shape routing and tool dispatch is the primitive no other paradigm provides — and the one that mechanically carries the lift at 70B.} The bottom-row primitive in Table~\ref{tab:comparison}, present only in the lightweight instantiation, replaces the structured-state machinery with a deterministic shape classifier feeding a small Python tool registry (Algorithm~\ref{alg:lightweight}); this is the Lightweight ReFlect (code-routed) breakout (\S\ref{sec:heavyweight-vs-lightweight}) and the foundation of the Full ReFlect headline result (\S\ref{sec:main-results}).

\begin{table}[h]
\caption{Comparison of reasoning paradigms. Rows describe primitives a paradigm provides. ReFlect cells annotated \cmark\textsuperscript{H} are present only in the heavyweight instantiation (\S\ref{sec:heavyweight}); \cmark\textsuperscript{L} are present only in the lightweight instantiation (\S\ref{sec:lightweight}, the system actually evaluated as Full ReFlect); unmarked \cmark are present in both. The lightweight design deliberately omits structured state, operators, and regime switching, replacing them with deterministic shape routing and tool dispatch (the bottom row), the primitive no other paradigm provides.}
\label{tab:comparison}
\centering
\small
\setlength{\tabcolsep}{3pt}
\begin{tabular}{@{}lccccccc@{}}
\toprule
& \textbf{CoT} & \textbf{ReAct} & \textbf{ToT} & \textbf{Self-Ref.} & \textbf{Reflexion} & \textbf{IterRes.} & \textbf{ReFlect} \\
\midrule
Structured state              & \xmark & \xmark & \xmark & \xmark & \xmark & \xmark\textsuperscript{$\dagger$} & \cmark\textsuperscript{H} \\
Separate inspection            & \xmark & \xmark & \xmark & \cmark & \cmark & \xmark & \cmark \\
Mid-task intervention          & \xmark & \xmark & \xmark & \xmark\textsuperscript{$*$} & \xmark & \xmark & \cmark \\
Assumption tracking            & \xmark & \xmark & \xmark & \xmark & \xmark & \xmark & \cmark\textsuperscript{H} \\
Rollback / checkpoint          & \xmark & \xmark & \xmark & \xmark & Restart & \xmark & \cmark\textsuperscript{H} \\
Branching / search             & \xmark & \xmark & \cmark & \xmark & \xmark & \xmark & \cmark\textsuperscript{H} \\
Regime switching               & \xmark & \xmark & \xmark & \xmark & \xmark & \xmark & \cmark\textsuperscript{H} \\
Bounded context                & \xmark & \xmark & \xmark & \xmark & \xmark & \cmark & \cmark \\
Shape routing \& tool dispatch & \xmark & \xmark & \xmark & \xmark & \xmark & \xmark & \cmark\textsuperscript{L} \\
No training required           & \cmark & \cmark & \cmark & \cmark & \cmark & \xmark\textsuperscript{$\ddagger$} & \cmark \\
\bottomrule
\end{tabular}

\vspace{2pt}
{\footnotesize $\dagger$\,IterResearch maintains bounded state as unstructured free text. $*$\,Self-Refine is post-output; in principle applicable mid-task but at prohibitive overhead. $\ddagger$\,IterResearch also works as a prompting strategy without training.}
\end{table}

\section{Execution trace example (lightweight harness on AIME)}
\label{app:trace}

To make Lightweight ReFlect (code-routed)'s control flow concrete, we walk through a single run on 2022 AIME~I Problem~13 (Qwen2.5-72B, sample index 15 of the AIME split). The problem asks for $N \bmod 1000$, where $N$ is the number of distinct numerators obtained when every repeating decimal of the form $0.\overline{abcd}$ is written as a reduced fraction; the ground-truth answer is $\mathbf{392}$. Phase tags in brackets refer to Lightweight ReFlect (code-routed)'s two active phases: $[\textsc{Execute}]$ (generate--parse--execute code) and $[\textsc{Consolidate}]$ (modal vote / stable-answer stop).

\begin{enumerate}[label=\textbf{S\arabic*.},leftmargin=*,itemsep=2pt]

\item $[\textsc{Execute}]$ Lightweight ReFlect (code-routed)'s domain router classifies the problem as \texttt{code\_solve} (AIME), so it enters the code phase with sample budget $K_\text{code}=3$ and draws three independent Python completions in parallel.

\item $[\textsc{Execute}]$ Each completion is parsed from its \verb|```python ... ```| fence and executed in a sandbox. Two of the three implement the brute-force enumeration \texttt{\{Fraction(n,9999).numerator\} for n in range(1,10000)\}}, printing \texttt{392}; the third mishandles the reducing step and prints \texttt{776}.

\item $[\textsc{Consolidate}]$ Lightweight ReFlect (code-routed) holds \texttt{code\_results=['392','776','392']}. The modal vote (\texttt{Counter.most\_common(1)}) selects $\mathbf{392}$ with $2/3$ agreement.

\item $[\textsc{Consolidate}]$ Because the vote returned a non-null answer, Lightweight ReFlect (code-routed) sets \texttt{finish\_reason=code\_solved} and terminates after a single outer step (\texttt{n\_steps=1}, \texttt{n\_calls=3}). The CoT self-consistency pool (\texttt{sc\_candidates}) and the stable-answer stop (\texttt{stable\_answer\_steps}) are fallback mechanisms that never engage here. Total reasoning tokens: 1{,}272.

\end{enumerate}

\noindent\textit{What the baselines did on the same problem.} Under identical settings, Direct, ReAct, and Self-Refine on Qwen2.5-72B all produced the incorrect answer $\mathbf{0}$ by computing $\varphi(9999)=6000$ and reducing modulo 1000 (the closed-form trap that counts only numerators coprime with 9999 while missing the non-coprime orbits). Reflexion returned no parseable final answer across three self-critique episodes. Lightweight ReFlect (code-routed)'s correctness here does not come from deeper symbolic reasoning: it comes from \emph{offloading} a number-theoretic enumeration to a Python sandbox and using $K=3$ modal voting to absorb one buggy code draft.

\section{Per-tool analysis}
\label{app:per-tool}

\begin{figure}[h]
\centering
\includegraphics[width=0.98\linewidth]{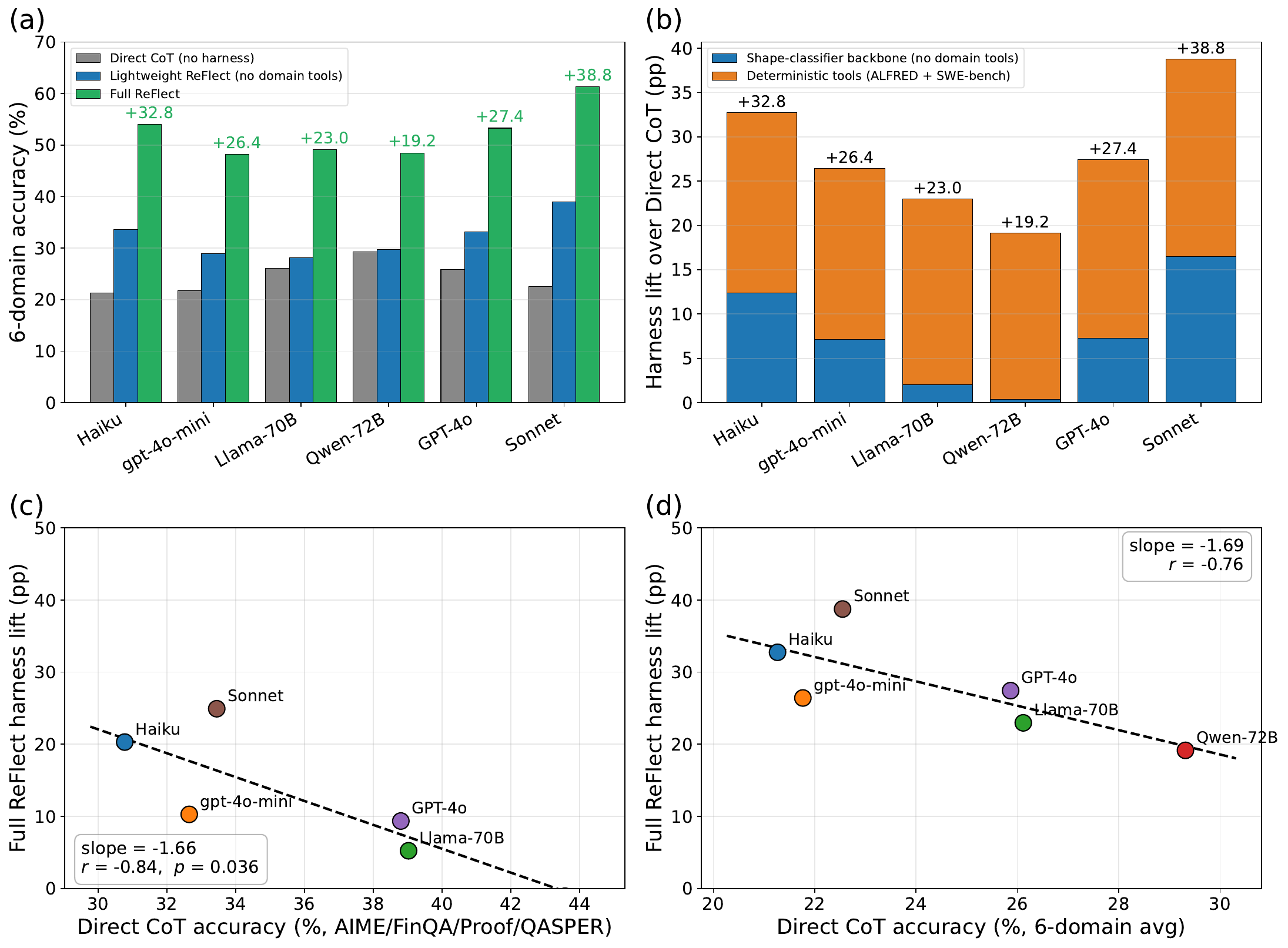}
\caption{Capability-compensation evidence across six models. \textbf{(a)} 6-domain accuracy across Direct CoT, Lightweight ReFlect (no domain tools), and Full ReFlect; numeric annotations are per-model lift over Direct. \textbf{(b)} Per-model decomposition of the lift over Direct CoT into the shape-classifier backbone (Lightweight ReFlect, no domain tools) and the deterministic tool layer (ALFRED state tracker $+$ SWE-bench diff verifier); the tool-layer contribution is near-uniform ($+19$ to $+22$~pp) while the backbone contribution is largest on the weakest models. \textbf{(c)} Full ReFlect harness lift vs Direct-CoT accuracy on the four LLM-driven domains (AIME, FinQA, ProofWriter, QASPER; excluding ALFRED and SWE-bench): slope $-1.66$ ($r=-0.84$, $p=0.036$). \textbf{(d)} Same fit across all 6 domains: slope $-1.69$ ($r=-0.76$).}
\label{fig:capability-ladder}
\label{fig:slope-refit}
\end{figure}

\begin{table}[h]
\caption{The seven computational shapes of the lightweight harness and their tools.}
\label{tab:shapes}
\centering
\small
\setlength{\tabcolsep}{5pt}
\begin{tabular}{@{}l l >{\raggedright\arraybackslash}p{8.0cm}@{}}
\toprule
\textbf{Shape} & \textbf{Tool} & \textbf{Mechanism} \\
\midrule
\textsc{Symbolic}    & \texttt{python\_symbolic}        & Python code (sympy/math/fractions), $K{=}3$ executions, modal vote \\
\textsc{Tabular}     & \texttt{python\_tabular}         & Python code (sympy/math/fractions), $K{=}3$ executions, modal vote \\
\textsc{Logical}     & \texttt{forward\_chain}          & Pure-Python Horn-clause forward chaining over extracted rules \\
\textsc{Evidence}    & \texttt{retrieval\_grounded}     & TF-IDF retrieval $\to$ grounded extraction ($K{=}1$) \\
\textsc{Procedural}  & \texttt{alfred\_state\_tracker}  & LLM proposes $K{=}5$ action sequences; Python preconditions model prefix-scores each \\
\textsc{Artifact}    & \texttt{diff\_verifier}          & LLM emits diff; harness validates format, retries on failure \\
\textsc{Fallback}    & \texttt{direct\_cot\_sc}         & $K{=}5$ direct CoT, modal vote \\
\bottomrule
\end{tabular}
\end{table}

Per-tool contribution within Full ReFlect, across the six evaluated models:

\paragraph{Forward-chain logic engine.} On the 9/50 ProofWriter problems where the deterministic forward-chain procedure commits to True/False, it achieves 100\% precision across all six models. On the 27-problem Unknown-delegation path the LLM handles unresolved cases, with performance ranging from 48\% (gpt-4o-mini) to 100\% (Claude Sonnet 4.5). Combined with the remaining 14 LLM-only fallback problems, overall ProofWriter accuracy ranges from 58\% to 96\%.

\paragraph{ALFRED state tracker.} A 212-line pure-Python verifier validates action-sequence preconditions without a simulator. Delivers 34--49\% across models, with the harness invoking up to 2 retries when proposed actions violate preconditions. The deterministic check is model-agnostic; LLM quality affects only the proposal step.

\paragraph{SymPy + tabular Python.} Code-generation tools dispatch each \textsc{Symbolic}-shaped problem (AIME) and \textsc{Tabular}-shaped problem (FinQA) to a 5-second-timeout sandbox, drawing $K=3$ samples and returning the modal vote. SymPy works for 18--46\% of AIME (model-dependent: gpt-4o-mini 18\%, Sonnet 46\%); Python tabular works for 74--84\% of FinQA across all models (saturated, model-independent within the experimental range).

\paragraph{Retrieval-grounded extractor.} TF-IDF retrieval over QASPER paper sections feeds the LLM a single grounded context. QASPER accuracy 6 to 13\% across models; the bottleneck is extraction (the LLM cannot pull span-level answers from retrieved chunks), not retrieval quality. Substituting OpenAI embeddings for TF-IDF moved gpt-4o-mini's QASPER score by $-0.7$~pp, indicating that retrieval upgrades are scale-conditional and the lever is elsewhere.

\paragraph{Diff verifier.} Forces a unified-diff format for SWE-bench patches with up to 2 retries on parse failure. Lifts every model from 0\% (no method produces valid diffs by default) to 82--87\% structural quality. Does not test the patch (no test execution), so the score measures patch validity rather than bug-fix correctness.

\paragraph{SC fallback.} Generic $K=5$ self-consistency on whatever the shape classifier could not assign to a specialist tool. Used on $<$10\% of problems across the six datasets.

\paragraph{Per-tool sampling temperatures.} The lightweight harness uses per-tool temperatures rather than a single global temperature: \textsc{Symbolic} draws $K{=}3$ candidate code samples at $T{=}0.7$ with a retry hint at $T{=}0.5$; \textsc{Tabular} draws $K{=}3$ at $T{=}0.7$; \textsc{Logical} (LLM-fallback path for Unknown problems) draws $K{=}5$ at $T{=}0.7$; \textsc{Evidence} runs grounded extraction at $T{=}0.2$ ($K{=}1$); \textsc{Procedural} draws $K{=}5$ candidate action sequences at $T{=}0.7$ with retry at $T{=}0.5$; \textsc{Artifact} draws diff candidates at $T{=}0.4$ with up to 3 format-failure retries; \textsc{Fallback} draws $K{=}5$ SC at $T{=}0.7$. The pilot study (\S\ref{sec:pilot-setup}) uses a single $T{=}0.6$ across all calls; the per-tool variation in Full ReFlect is inherited from the Lightweight ReFlect (interim) / Lightweight ReFlect (no domain tools) prompt-engineering pipeline.

\section{SWE-bench scorer}
\label{app:swebench-scorer}

SWE-bench uses a tiered structural-quality scorer rather than semantic correctness (which requires test execution in per-repo containers):

\begin{itemize}[leftmargin=*,itemsep=1pt]
\item \textbf{0.0}: output is not diff-formatted.
\item \textbf{0.3}: valid unified-diff format but targets non-code files.
\item \textbf{0.6}: targets code files (\texttt{.py}, \texttt{.js}, etc.) but added lines fail \texttt{ast.parse()}.
\item \textbf{1.0}: targets code files and added Python lines parse successfully.
\end{itemize}

This scorer produces 0.0 for all prior methods (Lightweight ReFlect (code-routed) and below output prose, not diffs) and 82--87\% for Full ReFlect (whose diff verifier forces valid diff output). Cross-version comparisons are therefore consistent: the tiered scorer adds granularity only where diff-formatted output exists.

\section{Convergence and termination behavior}
\label{app:convergence}

Table~\ref{tab:convergence} reports convergence rate (the fraction of 300-problem runs that terminate with an answer rather than exhausting the token budget) across the headline variants. Lightweight ReFlect (code-routed) (Full ReFlect) is the only variant in the series to reach $\geq$97\% convergence on both models (Llama 100\%, Qwen 97\%), reflecting the combined effect of P3 (Qwen stable-answer forced-stop) and P2 (code path termination when \texttt{print(...)} succeeds). Heavyweight ReFlect (operators ablated) is notable at the opposite extreme: Qwen converges on only 43\% of problems, with 170/300 runs hitting the token budget, illustrating how removing all structure leaves the model without a reliable termination signal.

\begin{table}[h]
\caption{Convergence rate across headline variants (300 problems each). ``Converged'' = terminated with an answer; not \texttt{budget\_exhausted}. Lightweight ReFlect (code-routed)'s near-100\% convergence is attributable to code-execution termination plus Qwen stable-answer stop.}
\label{tab:convergence}
\centering
\small
\setlength{\tabcolsep}{4pt}
\begin{tabular}{@{}lcccc@{}}
\toprule
\textbf{Variant} & \textbf{Llama conv. (\%)} & \textbf{Llama exh.} & \textbf{Qwen conv. (\%)} & \textbf{Qwen exh.} \\
\midrule
L2-SC + verbal CHECK                      & 91  & 28  & 93 & 21  \\
L2 + code-execution CHECK                 & 97  &  9  & 89 & 33  \\
L2 + linter-feedback REFLECT              & 98  &  7  & 90 & 29  \\
L2-Cross (cross-model)                    & 90  & 31  & 90 & 31  \\
Heavyweight ReFlect (operators ablated)   & 85  & 45  & 43 & 170 \\
Heavyweight ReFlect (stable termination)  & 97  &  8  & 87 & 38  \\
Lightweight ReFlect (code-routed)         & 100 &  0  & 97 & 10  \\
\bottomrule
\end{tabular}
\end{table}

\section{Repeated-error recurrence}
\label{app:repeated-error}

Cross-seed analysis on 3-seed runs (Llama-3.3-70B and Qwen2.5-72B, 300 problems $\times$ 5 methods). A ``stable error'' is a problem scored wrong on all 3 seeds with the same wrong answer, indicating a systematic, reproducible failure rather than stochastic variance.

\begin{table}[h]
\caption{Stable-error rate: fraction of wrong problems that produce the same wrong answer across all 3 seeds. Higher = more systematic (deterministic) failures.}
\centering\small
\begin{tabular}{@{}lcc@{}}
\toprule
\textbf{Method} & \textbf{Llama stable-err} & \textbf{Qwen stable-err} \\
\midrule
Direct CoT      & 21.4\% & 21.5\% \\
ReAct           & 28.4\% & 14.5\% \\
Self-Refine     & 21.0\% & 19.8\% \\
Reflexion       & \textbf{8.7\%}  & \textbf{10.8\%} \\
ReFlect (Lightweight ReFlect (code-routed))  & 30.6\% & 25.3\% \\
\bottomrule
\end{tabular}
\end{table}

Lightweight ReFlect (code-routed) has the highest stable-error rate (Llama 30.6\%, Qwen 25.3\%) because its code-execution path is deterministic: the same SymPy code produces the same wrong answer on every seed. Reflexion has the lowest (Llama 8.7\%, Qwen 10.8\%) because its retry loop introduces answer-level variance. This highlights a trade-off: deterministic tools produce more consistent (but also more systematically wrong) answers than stochastic retry methods.

\section{Official baseline comparison}
\label{app:official-baselines}

We ran Self-Refine~\citep{madaan2023selfrefine} and Reflexion~\citep{shinn2023reflexion} using the authors' prompt templates from their released code repositories on the same 300 problems, to verify that our reimplemented baselines are faithful.

\begin{table}[h]
\caption{Official vs.\ reimplemented baseline accuracy (\%). Deltas are within $\pm$3~pp noise, confirming our reimplementations are faithful.}
\centering\small
\begin{tabular}{@{}llccc@{}}
\toprule
\textbf{Model} & \textbf{Method} & \textbf{Official} & \textbf{Reimpl.} & \textbf{$\Delta$} \\
\midrule
Llama-3.3-70B & Self-Refine & 26.9 & 27.7 & $-$0.8 \\
Llama-3.3-70B & Reflexion   & 29.1 & 28.6 & $+$0.4 \\
Qwen2.5-72B  & Self-Refine & 24.7 & 27.8 & $-$3.1 \\
Qwen2.5-72B  & Reflexion   & 27.1 & 26.9 & $+$0.2 \\
\bottomrule
\end{tabular}
\end{table}

\section{Contamination probe}
\label{app:contamination}

We probed training-set memorization using two methods: (1)~bits-per-token ratio comparing answer likelihood under original vs.\ lexically-paraphrased questions (Llama, via Together.ai echo+logprobs), and (2)~verbatim continuation overlap (both models). A problem is flagged as ``likely memorized'' if either probe fires. The probe covers three domains at N=10 per model: AIME, ProofWriter, and SWE-bench. QASPER, ALFRED, and FinQA were not probed because retrieval-grounded QA problems are difficult to paraphrase without breaking context alignment, and the procedural/tabular tasks are paraphrased automatically by their structured input formats, making memorization unlikely a priori.

\begin{table}[h]
\caption{Memorization probe results (\% of N=10 sampled problems flagged per domain-model pair). AIME on Llama shows elevated memorization (40\%), consistent with public math-competition archives appearing in training data. All other domain-model pairs are at or below 20\%.}
\centering\small
\begin{tabular}{@{}lccc@{}}
\toprule
\textbf{Model} & \textbf{AIME} & \textbf{ProofWriter} & \textbf{SWE-bench} \\
\midrule
Llama-3.3-70B & 40\% & 0\% & 20\% \\
Qwen2.5-72B  &  0\% & 10\% &  0\% \\
\bottomrule
\end{tabular}
\end{table}

Since the paper's claim rests on \emph{relative} lift (Full ReFlect $-$ Direct), and memorization benefits both methods equally, the elevated AIME rate does not affect the main findings.

\section{Complete 28-variant sweep}
\label{app:variant-sweep}

This appendix backs Figure~\ref{fig:rq3-ablation-summary} (\S\ref{sec:heavyweight-vs-lightweight}) with the per-family written summary below and the per-variant detail in Table~\ref{tab:variant-sweep}. Table~\ref{tab:variant-sweep} retains the internal \texttt{vX.Y} reproducibility codes in column 1 as anchors to the source notebooks; descriptive labels for each row are given in the body text and the family-summary paragraph below. The table covers every variant tested in the full ablation sweep on both 70B models.

\paragraph{Figure~\ref{fig:rq3-ablation-summary} x-axis tag glossary.} The 18 short tags on the x-axis of Figure~\ref{fig:rq3-ablation-summary} map to the descriptive labels and design choices given in Table~\ref{tab:variant-tags}. The L2-SC bar aggregates the 16-cell verbal Self-Consistency sub-sweep into one mean-accuracy bar (see the L2-SC aggregation paragraph below); every other bar corresponds to a single variant.

\begin{table}[h]
\caption{Figure~\ref{fig:rq3-ablation-summary} x-axis tag glossary. Tags are grouped by family and sorted left-to-right matching the figure's bar order. Internal \texttt{vX.Y} codes are kept in the rightmost column as a cross-reference to Table~\ref{tab:variant-sweep} and the source notebooks; the body text uses only the descriptive labels.}
\label{tab:variant-tags}
\centering
\footnotesize
\setlength{\tabcolsep}{3pt}
\begin{tabular}{@{}l >{\raggedright\arraybackslash}p{3.8cm} >{\raggedright\arraybackslash}p{5.0cm} c@{}}
\toprule
\textbf{Tag} & \textbf{Descriptive label} & \textbf{Design choice} & \textbf{Code} \\
\midrule
\multicolumn{4}{@{}l}{\textit{Level-2 verbal SC}} \\
L2-SC & L2 self-consistency sweep & 16-cell SC sub-sweep, mean-aggregated & v5.x \\
L2-Verbal & L2-SC + verbal CHECK & best non-agentic Level-2; SC$+$verbal critique & v6.5 \\
\midrule
\multicolumn{4}{@{}l}{\textit{Level-2 with tools}} \\
L2-Code & L2 + code-execution CHECK & external Python interpreter as verifier & v7.0 \\
L2-Linter & L2 + deterministic linter & rule-based verifier on output format & v7.1 \\
L2-V$+$R & L2 + verbal CHECK + rules & verbal critique conditioned on rules & v7.2 \\
L2-LintFB & L2 + linter-feedback REFLECT & linter writes critique fed back to LLM & v7.3 \\
L2-Oracle & L2 + ground-truth oracle & upper-bound reference (verifier sees answer) & v7.4 \\
L2-Filter & L2 + linter sample-filter & linter filters bad SC samples before vote & v7.5 \\
L2-L$\wedge$C & L2 + linter $\wedge$ code-CHECK & conjunctive verifier (linter AND code) & v7.6 \\
\midrule
\multicolumn{4}{@{}l}{\textit{Level-2 cross-model}} \\
L2-Cross & L2 + cross-model verifier & solver and verifier are different LLMs & v8.0 \\
\midrule
\multicolumn{4}{@{}l}{\textit{Heavyweight Level-3 (pilot scoring)}} \\
HW-Full & Heavyweight ReFlect (full design) & state $+$ 4 operators $+$ regime FSM & v9.1 \\
HW-Agnos & Heavyweight ReFlect (agnostic refactor) & dataset-name router swapped for agnostic shape classifier & v9.5 \\
HW-Bare & Heavyweight ReFlect (operators ablated) & operators removed, bare multi-step CoT & v9.2 \\
HW-Best & Heavyweight ReFlect (with stable termination) & operators ablated $+$ deterministic stop $+$ $K{=}3$ SC & v9.3 \\
HW-Code & Heavyweight ReFlect (code-routed extension) & structured-state machinery replaced with deterministic Python routing on AIME$/$FinQA & v9.4 \\
\midrule
\multicolumn{4}{@{}l}{\textit{Lightweight Level-3 (RQ2)}} \\
LW-Base & Lightweight ReFlect (no domain tools) & agnostic shape classifier $+$ generic tool registry & v9.5-fixed \\
LW$+$ALFRED & Lightweight ReFlect (interim) & adds ALFRED state tracker only (no SWE diff yet) & v9.6 \\
ReFlect & \textbf{Full ReFlect} & adds SWE diff verifier on top of LW$+$ALFRED (RQ2 headline) & v9.7 \\
\bottomrule
\end{tabular}
\end{table}

\paragraph{L2-SC aggregation (Figure~\ref{fig:rq3-ablation-summary}).} The leftmost \texttt{L2-SC} bar in Figure~\ref{fig:rq3-ablation-summary} aggregates 16 distinct non-agentic Self-Consistency sub-variants (different $K$ values, EQUIV thresholds, unanimous-skip policies, sample-filter rules, etc.) into a single mean-accuracy bar at 17.0\%/18.0\%; their per-variant accuracies all live in the same Level-2 ceiling band (Llama 14--20\%, Qwen 14--22\%), so plotting them individually adds no information. Every other bar in Figure~\ref{fig:rq3-ablation-summary} corresponds to a single variant. The Level-2 verbal-SC family (v5--v6.x) is non-agentic (prompt-level only); the Level-2-with-tools family (v7.x) adds code-execution, linters, oracles, and conjunctions; the cross-model variant (v8.0) introduces an independent verifier; the Heavyweight ReFlect family (v9.1--v9.3) progresses from the full Level-3 design through operator-ablation variants; the Lightweight ReFlect family (v9.4 onward) progresses from code-routed through Full ReFlect. The pattern (variants cluster tightly in the 14 to 21\% band across all Level-2 mechanisms, plateau at 13 to 18\% for the Heavyweight family, and jump to 25 to 29\% only when computation routing arrives in the lightweight family) supports the paper's central claim. The Level-2 24-variant grid establishes the verifier FP-rate invariance (\S\ref{sec:heavyweight-vs-lightweight}, App.~\ref{app:error-correction}).

\paragraph{Family-level summary.}
\label{app:rq3-family-summary}
Aggregating the 28 variants into five mechanism families surfaces three patterns. \textbf{Three Level-2 families flatten in the same band.} Level-2 prompt verifiers (verbal SC, 17 cells) span Llama 17.0--19.3\% / Qwen 18.0--21.7\%, peaking at L2-Verbal with a pair-mean of 20.5\%; Level-2 with tools (7 variants) span Llama 17.0--19.6\% / Qwen 19.2--21.3\%, peaking at L2-LintFB with 20.2\%; the cross-model variant (L2-Cross) gives Llama 17.8\% / Qwen 17.3\% (17.6\% pair-mean). All three families fall in the same narrow band; no Level-2 variant exceeds the L2-Verbal peak of 20.5\% on either model. \textbf{Heavyweight Level-3 progresses through the Table~\ref{tab:heavyweight-trajectory} sequence.} The five Heavyweight ReFlect variants in Figure~\ref{fig:rq3-ablation-summary} use Table~\ref{tab:heavyweight-trajectory}'s pilot-scoring values (\S\ref{sec:heavyweight-eval}): \textsc{HW-Full} (16.7\%/13.3\%), \textsc{HW-Agnos} (19.0\%/17.7\%), \textsc{HW-Bare} (20.0\%/13.5\%), \textsc{HW-Best} (20.2\%/17.2\%) — the structured-state arm plateaus at 15.0--18.7\% pair-mean, sitting essentially at the Level-2 ceiling and matching ordinary baselines. The fifth variant \textsc{HW-Code} (26.9\%/29.2\%) jumps to 28.0\% pair-mean by replacing the structured-state machinery with deterministic Python routing on AIME and FinQA, the bridge to RQ2's polished lightweight redesign. \textbf{Polished Lightweight Level-3 (RQ2) extends further.} The three polished Lightweight ReFlect variants on the 70B pair (Together.ai/OpenRouter, framework scoring) span Llama 28.2--49.1\% / Qwen 29.6--48.5\%, with Full ReFlect the best at 48.8\% pair-mean. The interim variant plateaus near no-domain-tools at 29.6\% pair-mean (an intermediate refinement that does not lift 70B accuracy by itself); the $+19$~pp jump from Lightweight ReFlect (no domain tools, 28.95\%) to Full ReFlect (48.8\%) comes from the ALFRED state tracker and the SWE-bench diff verifier added together in Full ReFlect.

\paragraph{Lightweight Level-3 progression detail.}
The progression within the Lightweight ReFlect family on the 70B pair is: \emph{code-routed} (vLLM bf16) Llama 25.4\% / Qwen 28.9\% — independent SC $+$ Python code routing (the lightweight breakout); \emph{agnostic, vLLM} (vLLM bf16) Llama 18.0\% / Qwen 16.0\% — dataset-agnostic refactor that regresses at 70B; \emph{no domain tools}$^{\dagger}$ (Together.ai/OpenRouter) Llama 28.2\% / Qwen 29.7\% — dataset-agnostic shape classifier $+$ tool registry, recovers and exceeds the code-routed variant; \emph{interim}$^{\dagger}$ (Together.ai/OpenRouter) Llama 29.7\% / Qwen 29.6\% — intermediate refinement of \emph{no domain tools} (per-domain ALFRED stays 0.7\%/1.3\%, SWE-bench stays at 0\%, confirming neither domain tool was activated yet); \textbf{Full ReFlect}$^{\dagger}$ (Together.ai/OpenRouter) Llama 49.1\% / Qwen 48.5\% — adds the ALFRED state tracker (driving Llama ALFRED 0.7\%~$\to$~34.3\%, Qwen 1.3\%~$\to$~39.2\%) and the SWE-bench diff verifier (driving structural quality 0\%~$\to$~83.2\%/81.6\%) on top of \emph{no domain tools}; this is the headline RQ2 result. \textbf{$^{\dagger}$Serving caveat:} the \emph{code-routed} and \emph{agnostic-vLLM} rows use vLLM bf16 on the 70B pair (apples-to-apples with the Level-2 and Heavyweight Level-3 families); \emph{no domain tools}, \emph{interim}, and Full ReFlect use Together.ai (Llama Turbo FP8) and OpenRouter (Qwen) — see Appendix~\ref{app:serving}.

\begin{table}[h]
\caption{Full 28-variant accuracy sweep on the 70B pair (300 problems each). Variant codes (\texttt{vX.Y}) are reproducibility anchors mapping to the source notebooks; the body text uses the descriptive labels given in the rightmost column. Bold: best in series for the apples-to-apples vLLM rows. v5--v6.x are non-agentic Level-2 (verbal SC family); v7.x are Level-2 with tools (code, linter, oracle, conjunction); v8.0 is the Level-2 cross-model variant; v9.1--v9.3 are the Heavyweight ReFlect family; v9.4--v9.7 are the Lightweight ReFlect family. $^{\dagger}$Lightweight ReFlect (no domain tools, interim, Full ReFlect) use Together.ai (Llama Turbo FP8) / OpenRouter (Qwen) serving rather than vLLM bf16; reported here for the post-code-routed lightweight progression context. Family-level written summary in \S\ref{app:rq3-family-summary} (above) and visualization in Figure~\ref{fig:rq3-ablation-summary}, body \S\ref{sec:heavyweight-vs-lightweight}.}
\label{tab:variant-sweep}
\centering
\small
\setlength{\tabcolsep}{4pt}
\begin{tabular}{@{}lcc>{\raggedright\arraybackslash}p{6.4cm}@{}}
\toprule
\textbf{Variant} & \textbf{Llama (\%)} & \textbf{Qwen (\%)} & \textbf{Description} \\
\midrule
v5.x (16 variants) & 14--20 & 14--22 & Exhaustive non-agentic SC search \\
v6.5 (best non-agentic) & 19.3 & 21.7 & SC(N=3) + EQUIV + unanimous skip \\
v7.0 & 19.4 & 20.9 & Code-execution CHECK \\
v7.1 & 17.0 & 19.9 & Deterministic linter \\
v7.2 & 19.4 & 20.4 & Rules in verbal CHECK \\
v7.3 & 19.6 & 20.9 & Linter-formatted REFLECT \\
v7.4 & 17.2 & 19.2 & Oracle + linter \\
v7.5 & 18.1 & 20.7 & Linter sample filter \\
v7.6 & 17.0 & 21.3 & Linter $\wedge$ code CHECK \\
v8.0 & 17.8 & 17.3 & Cross-model (gpt-4o-mini) \\
\midrule
v9.1 & 14.0 & 12.3 & \emph{Heavyweight ReFlect (full design):} bugfixed Level-3 \\
v9.2 & 16.8 & 17.5 & \emph{Heavyweight ReFlect (operators ablated):} bare multi-step CoT \\
v9.3 & 18.4 & 17.8 & \emph{Heavyweight ReFlect (with stable termination):} $+$ in-traj.\ SC, P0 terminate \\
\textbf{v9.4} & \textbf{25.4} & \textbf{28.9} & \emph{Lightweight ReFlect (code-routed):} + independent SC, code routing, stable-stop \\
v9.5 & 18.0 & 16.0 & \emph{Lightweight ReFlect (agnostic, vLLM):} regression at 70B \\
\midrule
v9.5-fixed$^{\dagger}$ & 28.2 & 29.7 & \emph{Lightweight ReFlect (no domain tools):} dataset-agnostic shape classifier + tool registry \\
v9.6$^{\dagger}$       & 29.7 & 29.6 & \emph{Lightweight ReFlect (interim):} intermediate refinement (no measurable 70B lift) \\
\textbf{v9.7$^{\dagger}$} & \textbf{49.1} & \textbf{48.5} & \textbf{\emph{Full ReFlect:}} + ALFRED state tracker + SWE diff verifier \\
\bottomrule
\end{tabular}
\end{table}

\section{Error-correction quality across verification mechanisms}
\label{app:error-correction}

\begin{figure}[h]
\centering
\includegraphics[width=0.85\linewidth]{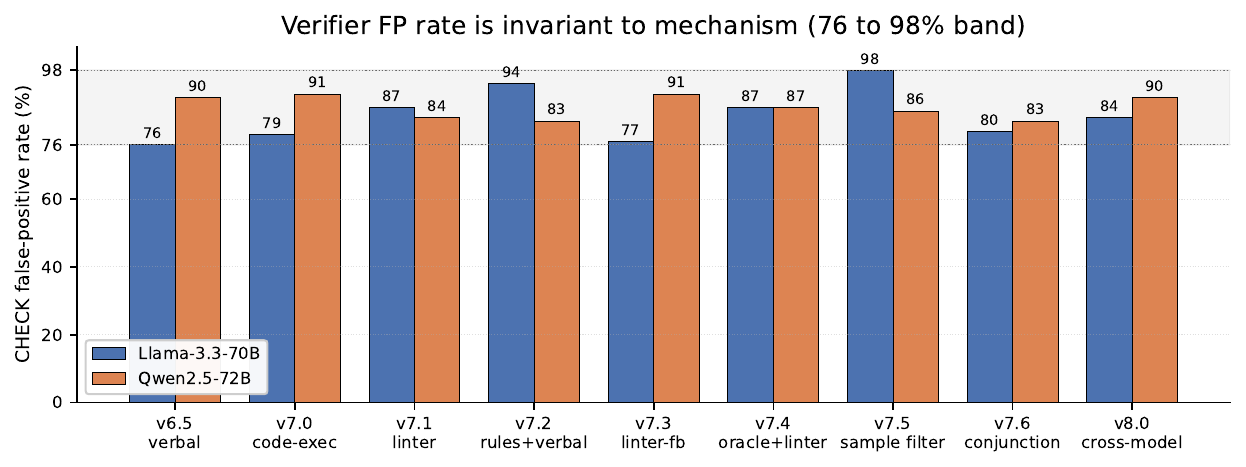}
\caption{Verifier false-positive rate across 9 mechanisms (Level-2 series: verbal, code-execution, linter, oracle, conjunction, cross-model) on Llama-3.3-70B and Qwen2.5-72B (300 problems per cell). FP rate stays in the 76 to 98\% band regardless of mechanism. Values in Table~\ref{tab:fp-rates}.}
\label{fig:fp-rate-invariance}
\end{figure}

Table~\ref{tab:fp-rates} presents CHECK false-positive rates across nine verification mechanisms tested in the Level-2 verifier series (verbal-SC, tool-augmented, and cross-model) on the full 300-problem benchmark, the empirical basis of the Level-2 ceiling claim in RQ3 (\S\ref{sec:heavyweight-vs-lightweight}). The false-positive (FP) rate is the fraction of problems where the CHECK module marked the answer \texttt{CORRECT} but the answer was actually wrong. Regardless of mechanism (verbal, code execution, deterministic linter, ground-truth oracle, conjunction of verifiers, or cross-model independent verification), the FP rate remains in the 76 to 98\% band. This invariance is the RQ3 negative result on prompt-level verification: the bottleneck is the verification task itself, not the mechanism or model. Figure~\ref{fig:fp-rate-invariance} visualizes the same data.

\begin{table}[h]
\caption{CHECK false-positive rate across verification mechanisms. FP = CHECK marked CORRECT but score $<$ 1.0. Values drawn from the Level-2 verifier series, v5--v8.0 (300 problems each). The cross-model variant (v8.0) uses gpt-4o-mini as an independent verifier that sees only the problem + answer, ruling out correlated errors as the explanation.}
\label{tab:fp-rates}
\centering
\small
\begin{tabular}{@{}lccc@{}}
\toprule
\textbf{CHECK mechanism} & \textbf{Llama FP (\%)} & \textbf{Qwen FP (\%)} & \textbf{Avg} \\
\midrule
L2-Verbal (verbal CHECK)                        & 76 & 90 & 83 \\
L2-Code (code-execution CHECK)                  & 79 & 91 & 85 \\
L2-Linter (deterministic linter)                & 87 & 84 & 86 \\
L2-V+R (rules + verbal)                         & 94 & 83 & 88 \\
L2-LintFB (linter-feedback REFLECT)             & 77 & 91 & 84 \\
L2-Oracle (oracle + linter)                     & 87 & 87 & 87 \\
L2-Filter (linter sample-filter)                & 98 & 86 & 92 \\
L2-L$\wedge$C (linter $\wedge$ code-CHECK)      & 80 & 83 & 82 \\
L2-Cross (gpt-4o-mini)                          & 84 & 90 & 87 \\
\bottomrule
\end{tabular}
\end{table}

Table~\ref{tab:reflect-success} presents the REFLECT success rate (the fraction of problems where the REFLECT module fired and produced a correct final answer) across five variants. Despite variations in feedback format (verbal critique, linter rules, code execution output), REFLECT success remains at 5 to 16\% across both models. This shows that the bottleneck is not feedback quality: when a 70B model's reasoning is wrong, providing feedback about \emph{what} is wrong does not enable it to find the \emph{right} answer.

\begin{table}[h]
\caption{REFLECT success rate: of problems where REFLECT fired, the percentage that produced a correct final answer. Includes fire count in parentheses. Data from the Level-2 verifier series (v5--v8.0) where REFLECT was enabled.}
\label{tab:reflect-success}
\centering
\small
\begin{tabular}{@{}lcccc@{}}
\toprule
\textbf{Variant} & \textbf{Llama fires} & \textbf{Llama succ.\ (\%)} & \textbf{Qwen fires} & \textbf{Qwen succ.\ (\%)} \\
\midrule
L2-Code     & 9  & 44.4 & 54 & 11.1 \\
L2-Linter   & 55 & 10.9 & 89 & 13.5 \\
L2-LintFB   & 9  & 22.2 & 54 & 11.1 \\
L2-L$\wedge$C & 63 & 15.9 & 118 & 16.3 \\
L2-Cross    & 77 & 15.9 & 59 & 12.9 \\
\bottomrule
\end{tabular}
\end{table}

\section{Systematic failures: full per-domain breakdown}
\label{app:systematic-detail}

The condensed prose in \S\ref{sec:heavyweight-vs-lightweight} omits per-model rescue/break detail and the headline visualization. Both are restored below.

\begin{figure}[h]
\centering
\includegraphics[width=0.95\linewidth]{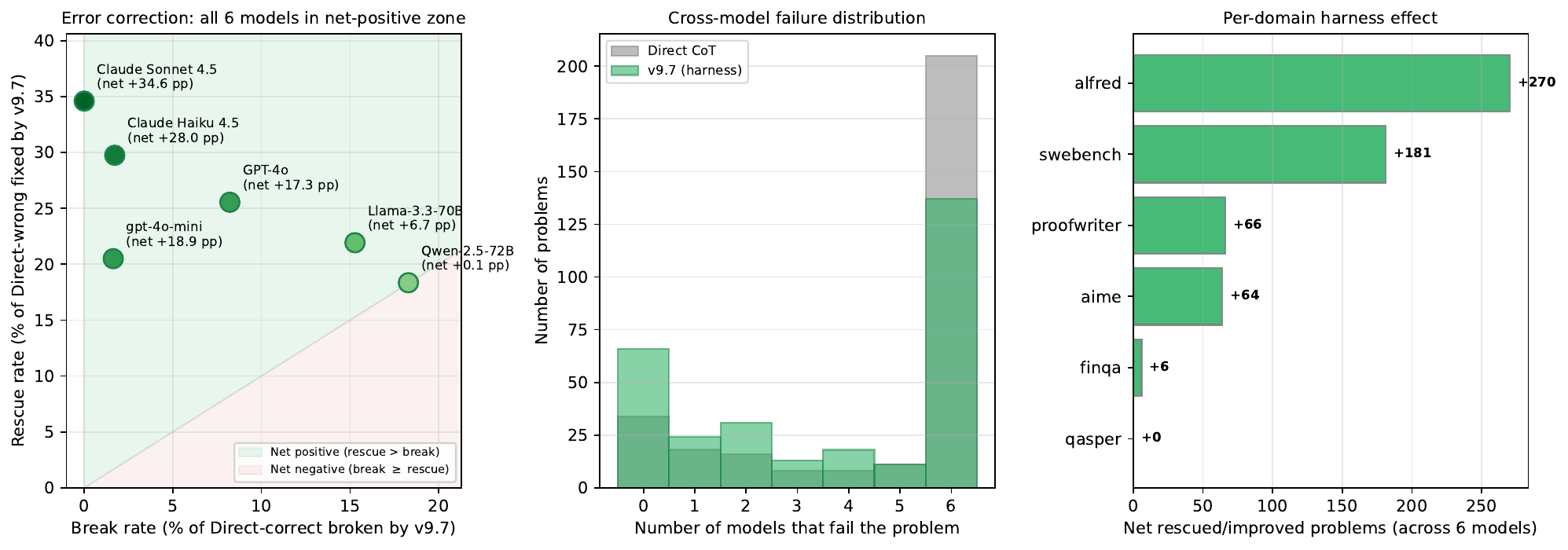}
\caption{Systematic-failure reduction across the six-model main grid. Left: rescue rate (Direct-wrong $\to$ Full ReFlect-right) vs.\ break rate (Direct-right $\to$ Full ReFlect-wrong) per model; all six models sit in the net-positive region. Center: histogram of how many models fail each problem; universal failures drop from 205 to 137. Right: per-domain net rescued problems (summed across 6 models); ALFRED ($+270$) and SWE-bench ($+181$) dominate, QASPER is net zero (extraction-capped).}
\label{fig:systematic-failures}
\end{figure}

\paragraph{Error-correction rate.} Across six models, Full ReFlect rescues a mean of 25.1\% of Direct's failures while breaking 7.5\% of its correct answers, a rescue-to-break ratio of 3.3:1. The best model (Claude Sonnet 4.5) achieves 34.6\% rescue with zero breakage. This confirms the harness is a net-positive intervention: it fixes substantially more than it damages.

\paragraph{Universal failures.} Under Direct CoT, 205 of 300 problems (68.3\%) are failed by \emph{all six models}. Under Full ReFlect, this drops to 137 (45.7\%): 68 problems that were universally impossible become solvable. Conversely, problems solved by all six models nearly doubles from 34 to 66. Of these, 28 problems flip from failure on $\geq$4 models (Direct) to success on $\geq$5 models (Full ReFlect): 6 AIME (SymPy makes them deterministically solvable), 6 ProofWriter (proof-chain forces correct reasoning), 15 SWE-bench (diff verifier produces valid patches), and 1 FinQA.

\paragraph{Per-domain effect.} ALFRED shows the strongest pattern: 90\% of problem-by-model pairs improve with zero regressions. This is the harness thesis in its purest form: a 200-line state tracker that verifies action preconditions \emph{systematically compensates} for what every LLM systematically gets wrong. QASPER is the control: with no tool addressing the extraction bottleneck, the harness shuffles errors without reducing them (31\% improved = 31\% worsened, net zero).

\section{Serving infrastructure}
\label{app:serving}

\paragraph{Cost computation (Table~\ref{tab:compute-matched}).} The \$/100~correct column applies a blended \$0.89/M-token rate (Together.ai \$0.88/M $+$ OpenRouter \$0.90/M, averaged) to the per-method token cost. \emph{Tokens/problem} is the mean total token count across the 300-problem 70B subset; \emph{Acc/1K~tokens} is accuracy (\%) divided by thousands of tokens consumed. Full ReFlect's lower token count vs.\ Lightweight ReFlect (code-routed) is mechanically attributable to its deterministic-Python tools: ALFRED state tracker (no LLM call), SWE-bench diff verifier (single call $+$ retry-as-code), QASPER grounded extraction (single call), and ProofWriter forward-chain (0 LLM tokens on 18\% of problems).

\begin{table}[h]
\caption{Compute-matched comparison on the 70B subset (Llama-3.3-70B and Qwen2.5-72B, 300 problems each, post-rescore FinQA scoring). Top block uses vLLM bf16 (Direct, ReAct, Self-Refine, Reflexion, code-routed are 3-seed averages; agnostic-vLLM is single seed). Bottom block ($^{\dagger}$) uses Together.ai (Llama FP8 Turbo) / OpenRouter (Qwen), single seed. Cost uses a blended \$0.89/M-token rate.}
\label{tab:compute-matched}
\centering
\small
\setlength{\tabcolsep}{4pt}
\begin{tabular}{@{}l r r r r@{}}
\toprule
\textbf{Method} & \textbf{Tokens/problem} & \textbf{Accuracy (\%)} & \textbf{Acc/1K tokens} & \textbf{\$/100 correct} \\
\midrule
Direct (1 run)                                                & 2,001  & 26.9 & 13.4 & 0.66 \\
ReAct (1 run)                                                 & 4,087  & 26.5 & 6.5 & 1.37 \\
Self-Refine (3 rounds)                                        & 8,361  & 27.2 & 3.2 & 2.74 \\
Reflexion (3 episodes)                                        & 32,062 & 27.8 & 0.9 & 10.26 \\
Lightweight ReFlect (code-routed)                             & 9,194  & 29.3 & 3.2 & 2.79 \\
Lightweight ReFlect (agnostic, vLLM)                          & 7,282  & 17.0 & 2.3 & 3.81 \\
\midrule
Lightweight ReFlect (no domain tools)$^{\dagger}$             & 2,939  & 28.9 & 9.8 & 0.91 \\
Full ReFlect$^{\dagger}$                                      & 1,993  & 48.8 & 24.5 & 0.36 \\
\bottomrule
\end{tabular}
\end{table}

The 70B pair is served via two distinct backends across the experiments, which a reproducer must match exactly to obtain the reported numbers:

\begin{itemize}[leftmargin=*,itemsep=1pt]
\item \textbf{vLLM~\citep{kwon2023vllm} (bf16, local GPUs).} Used for the motivating pilot (\S\ref{sec:motivation}, App. B--E) and the RQ3 ablation grid: the 24-variant Level-2 sweep and the full variant-sweep progression; the AIME execution-trace walkthrough (App.~\ref{app:trace}) is also reproduced on this backend. Models: \texttt{meta-llama/Llama-3.3-70B-Instruct} (bf16) and \texttt{Qwen/Qwen2.5-72B-Instruct} (bf16), tensor-parallel across 4 GPUs, max-model-len 32{,}768.
\item \textbf{Together.ai (Llama Turbo / FP8) and OpenRouter (Qwen).} Used for the 6-model main RQ2 lightweight results (Table~\ref{tab:main-results}), the capability ladder (\S\ref{sec:capability-ladder}, Figure~\ref{fig:capability-ladder}), the RQ2 compute-matched 3-seed comparison (Table~\ref{tab:compute-matched}), the official-baseline reimplementations (App.~\ref{app:official-baselines}), the contamination probe (App.~\ref{app:contamination}), and the seed-variance / repeated-error analyses (App.~\ref{app:repeated-error}). Models: \texttt{meta-llama/Llama-3.3-70B-Instruct-Turbo} (vendor-quantized FP8) and \texttt{qwen/qwen-2.5-72b-instruct}.
\end{itemize}

The Llama Turbo variant is FP8-quantized for serving efficiency and is numerically distinct from the bf16 vLLM-served Llama, so pilot and ablation numbers (App. B--E and App.~\ref{app:variant-sweep}) reproduce only on vLLM, while main-results numbers (Table~\ref{tab:main-results}) reproduce only on Together.ai. The 4 frontier API models are served as: Claude Haiku 4.5 (\texttt{claude-haiku-4-5}) and Claude Sonnet 4.5 (\texttt{claude-sonnet-4-5}) via the Anthropic API; gpt-4o-mini and GPT-4o via the OpenAI API. All API and vLLM calls share the same per-tool sampling parameters defined for Full ReFlect (\S\ref{sec:lightweight}); the pilot study (\S\ref{sec:pilot-setup}) uses a single temperature of 0.6 and top-$p$ 0.95 across all calls.

\section{Discussion: complementarity and meta-harness extension}
\label{app:discussion-extra}

The condensed §\ref{sec:discussion} omits per-paragraph detail on positioning, complementarity with prior work, and the meta-harness extension. Restored here.

\paragraph{Positioning (full).}
ReFlect is a harness system rather than a model modification. We define two harness instantiations (a lightweight one with shape routing and a tool registry, and a heavyweight one with structured state, operators, and regimes) and evaluate the lightweight instantiation across six heterogeneous domains with six backbone models, with no dataset-name routing in the framework code. The core contribution is the reframing: structural harnessing is not a capability enhancer added to the model, but a wrapper that converts open-ended failure into structurally-bounded computation. Memory, verification, and branching are instantiations of this harness principle (active failure-space reduction), not ad hoc modules.

\paragraph{Complementarity with existing work.}
ReFlect and IterResearch~\citep{chen2026iterresearch} operate at different levels of the reasoning stack and are naturally composable: IterResearch handles \emph{information gathering} via workspace reconstruction over hundreds of turns, while ReFlect handles \emph{reasoning quality} via shape-routed harness intervention. In a combined architecture, IterResearch would serve as the information-gathering substrate (Level 2), and ReFlect would add a structural harness layer (Level 3) that classifies each synthesized chunk to a computational shape, dispatches it to the appropriate verifier or extractor, and retries on structural failure. Notably, IterResearch reports substantial gains from iterative workspace reconstruction, but never analyzes \emph{whether the report correctly synthesizes the findings} or \emph{whether the reasoning built on the report is sound}, precisely the questions a harness layer can answer.

Conversely, ReFlect does not address capabilities where IterResearch excels: interaction scaling to thousands of turns, tight tool-use integration (web search, browser, code execution), and trained efficiency via reinforcement learning. For tasks requiring extensive information gathering, IterResearch's approach is better suited; for tasks requiring multi-step reasoning with structural validation and retry, ReFlect is the right system. Similarly, ReFlect's harness could augment Reflexion's~\citep{shinn2023reflexion} cross-episode memory with within-episode shape routing and structural validation. The paradigm comparison (Table~\ref{tab:paradigm-comparison}) is not a replacement hierarchy but a capability stack.

\paragraph{Toward meta-level harness optimization.}
The current ReFlect instantiations use a manually designed harness: hand-built shape classifier, fixed tool registry, hand-tuned retry policy. A natural extension is a \emph{two-level architecture}: an inner-loop reflect-harness agent executes tasks under the structurally-bounded interface described above, while an outer-loop \emph{meta-harness} optimizes the harness configuration itself (shape boundaries, tool selection, retry budgets, format thresholds, fallback rules, and stop criteria) using execution traces, structural-failure logs, and multi-objective evaluation (task success vs.\ token cost vs.\ convergence speed). This outer loop treats the harness as an optimizable code object rather than a fixed scaffold, searching over harness designs using the same trajectory data that the inner loop produces. The meta-harness perspective transforms ReFlect from a single-configuration system into a design space amenable to systematic exploration.

\newpage
\section*{NeurIPS Paper Checklist}

\begin{enumerate}

\item {\bf Claims}
    \item[] Question: Do the main claims made in the abstract and introduction accurately reflect the paper's contributions and scope?
    \item[] Answer: \answerYes{}
    \item[] Justification: The abstract and \S\ref{sec:intro} introduce the harness reframing, three research questions (RQ1--RQ3), and five contributions; each is empirically substantiated in \S\ref{sec:method}--\S\ref{sec:heavyweight-vs-lightweight} (Tables~\ref{tab:heavyweight-trajectory}, \ref{tab:main-results} and Figure~\ref{fig:capability-ladder}). Headline numerical claims (41--56\% correctness across six models, $+7$ to $+29$~pp lift, slope $-1.69$ on six domains and $-1.66$ on the four LLM-driven domains, heavyweight structured-state arm 15.0--18.7\% pair-mean under pilot scoring with the in-family code-routed extension reaching 28.0\%) are verified cell-by-cell against per-problem CSVs released in \texttt{Resources/data/raw\_results/} and reproduced by \texttt{cost\_per\_correct.py}, \texttt{slope\_refit.py}, and \texttt{systematic\_failures.py}.

\item {\bf Limitations}
    \item[] Question: Does the paper discuss the limitations of the work performed by the authors?
    \item[] Answer: \answerYes{}
    \item[] Justification: \S\ref{sec:discussion} discusses scope and boundary conditions: (i) a base-capability prerequisite that the heavyweight design fails to meet at 70B scale (\S\ref{sec:heavyweight-diagnoses-section}: state-extraction yields zero evidence on 84\% of problems, operators fire on $\leq 5\%$ of steps); (ii) the dataset-agnostic router pairs with task-shaped specialist tools, so a domain whose bottleneck the harness does not yet target (e.g., QASPER span extraction at 6--13\%) is bounded by that capability rather than by the harness; (iii) the 50-instance-per-domain scope on the 6-model lightweight grid (the compute-matched 3-seed verification on the 70B subset, Appendix~\ref{app:serving}, mitigates seed sensitivity at the 70B scale).

\item {\bf Theory assumptions and proofs}
    \item[] Question: For each theoretical result, does the paper provide the full set of assumptions and a complete (and correct) proof?
    \item[] Answer: \answerNA{}
    \item[] Justification: The paper presents a paradigm-level reframing and an empirical evaluation. It contains no formal theorems, lemmas, or proofs. The capability-compensation slope analysis (Figure~\ref{fig:capability-ladder} panels c--d) is a linear regression reporting Pearson $r=-0.76$ on the 6-domain fit and $r=-0.84$, $p=0.036$ on the 4-domain refit (slope refit script \texttt{slope\_refit.py} in \texttt{Resources/code/}); these are empirical statistical claims rather than theoretical results.

    \item {\bf Experimental result reproducibility}
    \item[] Question: Does the paper fully disclose all the information needed to reproduce the main experimental results of the paper to the extent that it affects the main claims and/or conclusions of the paper (regardless of whether the code and data are provided or not)?
    \item[] Answer: \answerYes{}
    \item[] Justification: \S\ref{sec:setup} fully specifies the six benchmarks, six backbone models, all baseline methods (with episode/round counts), inference settings (pilot uses $T{=}0.6$, top-$p\,0.95$; per-tool sampling temperatures for Full ReFlect listed in Appendix~\ref{app:per-tool}), and per-method evaluation metrics. The seven specialized tools and shape classifier are described in \S\ref{sec:lightweight-section}; the heavyweight design (Algorithm~\ref{alg:reflect}) is in Appendix~\ref{app:heavyweight-detail}. The released \texttt{Resources/} package contains the canonical implementations: \texttt{reflect\_framework\_full.py} (Full ReFlect), \texttt{reflect\_framework\_heavyweight\_\{full,fix,bare,best\}.py} (four heavyweight variants), \texttt{reflect\_framework\_lightweight\_\{vllm,code,base,interim\}.py} (four lightweight variants), shared utilities (\texttt{reflect\_framework\_common.py}, \texttt{reflect\_state.py}, \texttt{api\_helpers.py}, \texttt{domain\_linters.py}), and three reproducer scripts (\texttt{cost\_per\_correct.py}, \texttt{slope\_refit.py}, \texttt{systematic\_failures.py}) that regenerate the headline numbers from per-problem CSVs in \texttt{data/raw\_results/}.

\item {\bf Open access to data and code}
    \item[] Question: Does the paper provide open access to the data and code, with sufficient instructions to faithfully reproduce the main experimental results, as described in supplemental material?
    \item[] Answer: \answerYes{}
    \item[] Justification: The supplementary \texttt{Resources/} package contains: (i) the Full ReFlect implementation (\texttt{code/reflect\_framework\_full.py}) and the eight intermediate variants used for the RQ1/RQ3 ablations; (ii) per-problem CSVs in \texttt{data/raw\_results/} organized by variant (\texttt{direct/}, \texttt{full\_reflect/}, \texttt{heavyweight\_\{full,fix,bare,best\}/}, \texttt{lightweight\_\{vllm,code,base,interim\}/}); (iii) aggregated analysis CSVs in \texttt{data/analysis/} (\texttt{capability\_ladder\_per\_domain.csv}, \texttt{capability\_ladder\_summary.csv}, \texttt{compute\_matched\_summary.csv}, \texttt{cost\_per\_correct.csv}, \texttt{rescue\_rate\_per\_\{model,domain\}.csv}); (iv) three reproducer scripts (\texttt{cost\_per\_correct.py}, \texttt{slope\_refit.py}, \texttt{systematic\_failures.py}); (v) \texttt{requirements.txt} and three \texttt{README.md} files (root, \texttt{code/}, \texttt{data/}). All six evaluation benchmarks (SWE-bench Lite, QASPER, ProofWriter depth-5, AIME 2022--2024, ALFRED, FinQA) are publicly available, cited to their canonical publications.

\item {\bf Experimental setting/details}
    \item[] Question: Does the paper specify all the training and test details (e.g., data splits, hyperparameters, how they were chosen, type of optimizer) necessary to understand the results?
    \item[] Answer: \answerYes{}
    \item[] Justification: \S\ref{sec:setup} specifies 50 instances per domain (300 total per model on the 6-model main grid), all six benchmarks, all six backbone models, baseline configurations (Self-Refine 3 rounds, Reflexion 3 episodes), and serving infrastructure (vLLM bf16 for the 70B pair on the pilot/RQ1/RQ3; Together.ai/OpenRouter for the 6-model main grid; Anthropic/OpenAI APIs for the four frontier models). Per-tool sampling temperatures, $K$ values, retry budgets, and the fallback policy for the seven shape-specific tools are in Appendix~\ref{app:per-tool} (the pilot uses a single $T{=}0.6$, top-$p\,0.95$). No training was performed: all methods are inference-time at fixed model checkpoints. The released harness modules in \texttt{Resources/code/} are the canonical specification of every prompt, sampling configuration, and retry policy.

\item {\bf Experiment statistical significance}
    \item[] Question: Does the paper report error bars suitably and correctly defined or other appropriate information about the statistical significance of the experiments?
    \item[] Answer: \answerYes{}
    \item[] Justification: The capability-compensation slope (Figure~\ref{fig:capability-ladder} panels c--d; reproduced by \texttt{slope\_refit.py} on \texttt{capability\_ladder\_per\_domain.csv}) is reported with Pearson $r=-0.76$ on the 6-domain fit and $r=-0.84$, $p=0.036$ on the 4-domain LLM-driven refit. The 70B compute-matched table (Table~\ref{tab:compute-matched}, Appendix~\ref{app:serving}) reports 3-seed averages for vLLM Direct CoT, ReAct, Self-Refine, Reflexion, and Lightweight ReFlect (code-routed), aggregated from \texttt{compute\_matched\_summary.csv}. The pilot reports Wilson 95\% confidence intervals on the course-correction rate (1/60 Qwen, 0/60 Llama; CI $\leq 8.9\%$). The 28-variant RQ3 ablation uses 300 problems per cell on the 70B pair, narrowing per-variant uncertainty.

\item {\bf Experiments compute resources}
    \item[] Question: For each experiment, does the paper provide sufficient information on the computer resources (type of compute workers, memory, time of execution) needed to reproduce the experiments?
    \item[] Answer: \answerYes{}
    \item[] Justification: Appendix~\ref{app:serving} (Serving infrastructure) details per-experiment compute: vLLM bf16 served locally on 4 GPUs with tensor parallelism and \texttt{max\_model\_len}~$=32{,}768$ (used for the pilot, RQ1 heavyweight, and RQ3 ablation grid); Together.ai (Llama-3.3-70B-Instruct-Turbo, FP8) and OpenRouter (Qwen2.5-72B-Instruct) for the 6-model main grid; Anthropic and OpenAI APIs for the four frontier models. Per-method per-problem token budgets (Table~\ref{tab:compute-matched}): Direct (2{,}001), ReAct (4{,}087), Self-Refine (8{,}361), Reflexion (32{,}062), Lightweight ReFlect code-routed (9{,}194), Lightweight ReFlect no-domain-tools (2{,}939), Full ReFlect (1{,}993). Cost is reported under a blended \$0.89/M-token rate; Full ReFlect achieves 48.8\% pair-mean accuracy at \$0.36 per 100 correct.

\item {\bf Code of ethics}
    \item[] Question: Does the research conducted in the paper conform, in every respect, with the NeurIPS Code of Ethics \url{https://neurips.cc/public/EthicsGuidelines}?
    \item[] Answer: \answerYes{}
    \item[] Justification: The research uses publicly released benchmarks and publicly accessible language models. No human subjects, no scraped or sensitive data, no model training, and no model release with elevated misuse risk. All cited assets are appropriately attributed. The authors have reviewed the NeurIPS Code of Ethics and confirm compliance.

\item {\bf Broader impacts}
    \item[] Question: Does the paper discuss both potential positive societal impacts and negative societal impacts of the work performed?
    \item[] Answer: \answerYes{}
    \item[] Justification: The paper proposes a deterministic harness wrapper around language models that disproportionately benefits weaker, smaller models (capability-compensation slope $-1.69$). Positive impact: democratizes complex-reasoning capability to users running smaller, more accessible models on lower compute budgets. Negative-impact considerations: harnesses that improve LLM reliability on coding/math tasks can enable misuse in any setting where reliable LLM output matters (e.g., automated decision-making in unintended deployments). The paradigm is model-agnostic and training-free, so it does not introduce new model-distribution risks beyond those of the underlying LLMs being wrapped.

\item {\bf Safeguards}
    \item[] Question: Does the paper describe safeguards that have been put in place for responsible release of data or models that have a high risk for misuse (e.g., pre-trained language models, image generators, or scraped datasets)?
    \item[] Answer: \answerNA{}
    \item[] Justification: The paper does not release a pre-trained language model, an image generator, or a scraped dataset. The released artifacts are inference-time framework code (a deterministic harness around existing public models) and result CSVs over public benchmarks. None of these have an elevated misuse profile beyond the underlying public assets they wrap.

\item {\bf Licenses for existing assets}
    \item[] Question: Are the creators or original owners of assets (e.g., code, data, models), used in the paper, properly credited and are the license and terms of use explicitly mentioned and properly respected?
    \item[] Answer: \answerYes{}
    \item[] Justification: All six benchmarks (SWE-bench Lite, QASPER, ProofWriter, AIME 2022--2024, ALFRED, FinQA) and all backbone models (Qwen2.5-72B-Instruct, Llama-3.3-70B-Instruct, Claude Haiku 4.5, gpt-4o-mini, GPT-4o, Claude Sonnet 4.5) are properly cited with their original publications (\citep{jimenez2024swebench, dasigi2021qasper, tafjord2021proofwriter, aime2024, shridhar2020alfred, chen2021finqa, qwen2024qwen25, llama2024llama33}). vLLM serving infrastructure is cited \citep{kwon2023vllm}. Each benchmark and model is used within its publicly released license terms; no proprietary or restricted asset is used.

\item {\bf New assets}
    \item[] Question: Are new assets introduced in the paper well documented and is the documentation provided alongside the assets?
    \item[] Answer: \answerYes{}
    \item[] Justification: New assets in the released \texttt{Resources/} package: 9 ReFlect harness modules (\texttt{reflect\_framework\_full.py}; \texttt{reflect\_framework\_heavyweight\_\{full,fix,bare,best\}.py}; \texttt{reflect\_framework\_lightweight\_\{vllm,code,base,interim\}.py}); shared utilities (\texttt{reflect\_framework\_common.py}, \texttt{reflect\_state.py}, \texttt{api\_helpers.py} including the SWE-bench tiered scorer, \texttt{domain\_linters.py}); 3 reproducer scripts (\texttt{cost\_per\_correct.py}, \texttt{slope\_refit.py}, \texttt{systematic\_failures.py}); per-problem CSVs in \texttt{data/raw\_results/} (60+ files spanning Direct CoT and 9 ReFlect variants); 5 aggregated analysis CSVs. Each subdirectory carries a \texttt{README.md} (root, \texttt{code/}, \texttt{data/}) describing module/CSV semantics; \texttt{requirements.txt} pins dependencies. The heavyweight design is documented as Algorithm~\ref{alg:reflect} in Appendix~\ref{app:heavyweight-detail}. All artifacts are anonymized for double-blind review.

\item {\bf Crowdsourcing and research with human subjects}
    \item[] Question: For crowdsourcing experiments and research with human subjects, does the paper include the full text of instructions given to participants and screenshots, if applicable, as well as details about compensation (if any)?
    \item[] Answer: \answerNA{}
    \item[] Justification: The research does not involve crowdsourcing or human subjects. All evaluation is on publicly released benchmarks; all model outputs are generated programmatically.

\item {\bf Institutional review board (IRB) approvals or equivalent for research with human subjects}
    \item[] Question: Does the paper describe potential risks incurred by study participants, whether such risks were disclosed to the subjects, and whether Institutional Review Board (IRB) approvals (or an equivalent approval/review based on the requirements of your country or institution) were obtained?
    \item[] Answer: \answerNA{}
    \item[] Justification: The research does not involve human subjects, so no IRB approval is required.

\item {\bf Declaration of LLM usage}
    \item[] Question: Does the paper describe the usage of LLMs if it is an important, original, or non-standard component of the core methods in this research? Note that if the LLM is used only for writing, editing, or formatting purposes and does \emph{not} impact the core methodology, scientific rigor, or originality of the research, declaration is not required.
    \item[] Answer: \answerYes{}
    \item[] Justification: LLMs are central to the research as the experimental subjects evaluated under the proposed harness. Six LLMs (Qwen2.5-72B-Instruct, Llama-3.3-70B-Instruct, Claude Haiku 4.5, gpt-4o-mini, GPT-4o, Claude Sonnet 4.5) are explicitly named in \S\ref{sec:setup} and used as the underlying reasoning substrate inside the harness. The harness wraps the LLM in a deterministic shape-classification + tool-dispatch loop; the LLM's role is candidate generation inside structurally-bounded slots. The Acknowledgments section explicitly declares that LLMs were used \emph{only} for grammar and style checking of the manuscript text and were not used for ideation, experimental design, code generation, results analysis, or content authorship.

\end{enumerate}

\end{document}